%% file: main.tex
\documentclass[10pt,twocolumn,letterpaper]{article}

\usepackage[pagenumbers]{cvpr} 

\input{preamble}
\definecolor{cvprblue}{rgb}{0.21,0.49,0.74}
\usepackage[pagebackref,breaklinks,colorlinks,allcolors=cvprblue]{hyperref}

\usepackage{multirow}

\def\paperID{****} 
\def\confName{CVPR}
\def\confYear{2026}

\title{FHAvatar: Fast and High-Fidelity Reconstruction of Face-and-Hair Composable 3D Head Avatar from Few Casual Captures}

\def\paperName{FHAvatar }
\def\paperNameWOSpace{FHAvatar}

\author{
Yujie Sun\textsuperscript{1,2}\footnotemark[1], \quad
Zhuoqiang Cai\textsuperscript{1,2}\footnotemark[1], \quad
Chaoyue Niu\textsuperscript{1}\footnotemark[2], \quad
Jianchuan Chen\textsuperscript{2}, \\
Zhiwen Chen\textsuperscript{2}, \quad
Chengfei Lv\textsuperscript{2}, \quad
Fan Wu\textsuperscript{1} \\
\textsuperscript{1}Shanghai Jiao Tong University \quad
\textsuperscript{2}Alibaba Group \\
}

\begin{document}

\twocolumn[
\maketitle
\vspace{-3em}
\input{figs/teaser}
\vspace{-0.2em}
\bigbreak
]

\renewcommand{\thefootnote}{\fnsymbol{footnote}}
\footnotetext[1]{Intern at Alibaba Group. $^\dag$Corresponding author.}

\input{sec/0_abstract}    
\input{sec/1_intro}
\input{sec/2_related_works}
\input{sec/4_method}
\input{sec/5_experiments}
\input{sec/6_conclusion}

\section*{Acknowledgements}

This work was supported in part by National Key R\&D Program of China (No. 2023YFB4502400), in part by China NSF grant (No. 62572299, No. 62441236, No. 62372296, No. 62432007, No. U24A20326, No. U25A6024, No. U25A20437), in part by Fundamental and Interdisciplinary Disciplines Breakthrough Plan of the Ministry of Education of China (No. JYB2025XDXM103), in part by Alibaba Innovation Research (AIR) Program, Tencent WeChat Research Program, and SJTU-Huawei Explore X Gift Fund.

Chaoyue Niu is the corresponding author.

{
    \small
    \bibliographystyle{ieeenat_fullname}
    \bibliography{main}
}

\input{sec/X_suppl}

\end{document}

%% file: figs/teaser.tex
\begin{center}
    \includegraphics[width=1\linewidth]{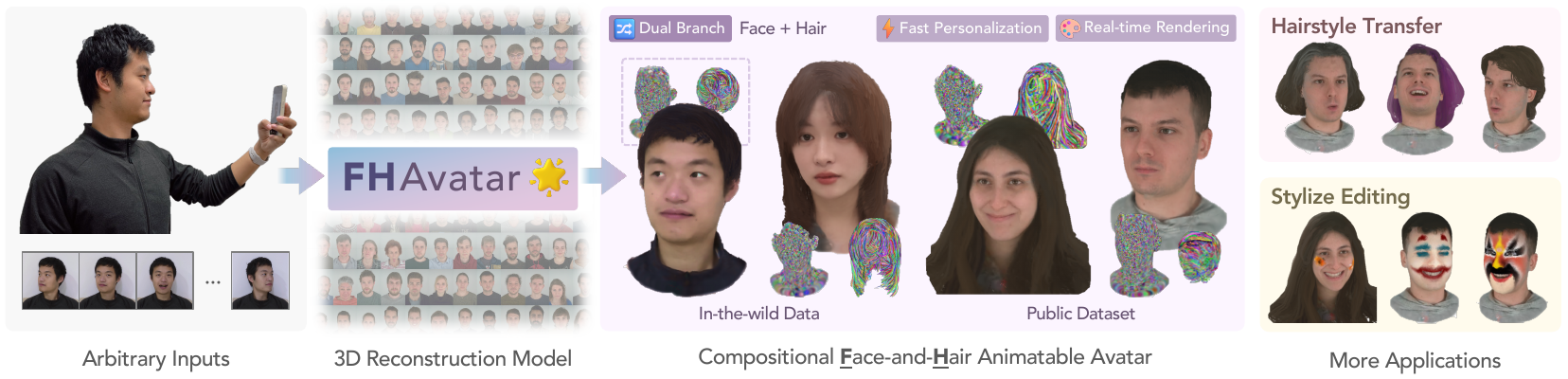}
    \end{center} \vspace{-1.6em} 
    \captionof{figure}{\textbf{\paperName} generates high-fidelity, animatable 3D head avatars from arbitrary inputs, such as a few phone shots, in just minutes. The reconstructed heads contain composable face and hair components, enabling hairstyle transfer and convenient stylized editing.}
\label{fig:teaser}

%% file: sec/0_abstract.tex
\begin{abstract}

We present \paperNameWOSpace{}, a novel framework for reconstructing 3D Gaussian avatars with composable face and hair components from an arbitrary number of views.
Unlike previous approaches that couple facial and hair representations within a unified modeling process, we explicitly decouple two components in texture space by representing the face with planar Gaussians and the hair with strand-based Gaussians. To overcome the limitations of existing methods that rely on dense multi-view captures or costly per-identity optimization, we propose an aggregated transformer backbone to learn geometry-aware cross-view priors and head-hair structural coherence from multi-view datasets, enabling effective and efficient feature extraction and fusion from few casual captures. Extensive quantitative and qualitative experiments demonstrate that \paperNameWOSpace{} achieves state-of-the-art reconstruction quality from only a few observations of new identities within minutes, while supporting real-time animation, convenient hairstyle transfer, and stylized editing, broadening the accessibility and applicability of digital avatar creation.
\end{abstract}

%% file: sec/1_intro.tex
\section{Introduction}
\label{sec:intro}

The reconstruction of photorealistic and animatable 3D head avatars has broad applications in diverse scenarios, including online communication, interactive gaming, and virtual streaming. Recent advances~\cite{fg2023flame-in-nerf, cvpr2023insta, cvpr2024flashavatar, cvpr2024gaussianavatars, tvcg2025gaussianhead} have achieved remarkable progress by leveraging powerful 3D representations such as Neural Radiance Fields (NeRF)~\cite{eccv2020nerf} and 3D Gaussian Splatting (3DGS)~\cite{tog20233dgaussian}, in conjunction with parametric geometric models, enhancing both modeling and animation quality. 

However, existing work often overlook the intrinsic differences between facial and hair regions. While facial geometry exhibits relatively consistent structural priors across identities, hair varies dramatically in style, density, and length. Many current models~\cite{cvpr2024gaussianavatars, cvpr2024flashavatar, cvpr2025rgbavatar} treat hair as an extension or deformation of the scalp, neglecting the fine-grained strand-level geometry that defines realistic appearance. This simplification often leads to degraded visual quality and limits downstream applications such as hairstyle editing, transfer, and physical simulation. High-fidelity hair modeling at the strand level typically requires manual intervention, complex capture setups, or costly optimization.

Beyond the lack of differentiation between facial and hair structures, many high-quality head reconstruction methods~\cite{cvpr2024gaussianavatars, cvpr2024flashavatar, siggragh2024monogaussianavatar, cvpr2024spalttingavatar, cvpr2025mega} remain identity-specific. They rely on dense multi-view data and long acquisition or processing times, which greatly restrict their usability in practical, consumer-level scenarios. To alleviate these constraints, several recent approaches~\cite{nips2024gagavatar, corr2025lam, corr2025avat3r, cvpr2025gasp} propose to train generalized feed-forward models on large video datasets~\cite{cvpr2022vfhq, tog2023nersemble, nips2024codecavatarstudio}. While promising, these methods are typically limited to single-view or fixed-viewpoint inputs, lacking the flexibility needed for casual mobile capture. Moreover, these methods still struggle with robustness under sparse input images and tend to introduce 3D inconsistency artifacts during reconstruction.

In this work, we propose \textbf{\paperNameWOSpace}, a novel framework for reconstructing animatable 3D Gaussian head avatars with composable face and hair components from only a few images captured by mobile device, completing the entire process within minutes.

\paperNameWOSpace{} adopts an encoder–decoder architecture specifically designed for disentangled face-hair reconstruction. From the input images, region-specific features are extracted: coarse-grained information on hair style, length, and density, as well as the face region. A dual-branch decoder independently reconstructs the face and hair. The decoded hair Gaussians are modeled as a series of connected strand Gaussians bound to the root of the mesh, aligning with the physical characteristics of hair. To balance efficiency and quality, we predict a density map, which allows for the random sampling of strands and adaptive adjustment of strand parameters according to hair length, reducing the number of Gaussians while maintaining visual quality. 

Between the encoder and the decoder is an aggregated transformer backbone, inspired by recent advances in scene-understanding Vision Transformers~\cite{corr2025vggt, corr2025pi3}. This design efficiently aggregates multi-view features and supports a variable number of input images, enabling the model to learn priors from both monocular and multi-view data, while maintaining strong generalization and high reconstruction quality. Following recent GAN-based generative models~\cite{siggraph2024gghead, icassp2025avatargan}, geometry and texture representations are learned in the texture space and are decoded into 3D Gaussians, which are then mapped onto a geometric mesh to quickly produce an animatable avatar.

Trained on a large-scale multi-view dataset, \paperNameWOSpace{} can reconstruct unseen identities in a single forward pass, followed by an optional lightweight refinement step applied only to the decoders, all within a few minutes. The resulting avatars support real-time animation and exhibit superior quality in hairstyle transfer and stylized face editing tasks.

In summary, our contributions are as follows:
\begin{itemize}
    \item We present \textbf{\paperNameWOSpace}, a new pipeline for generating animatable, face-and-hair compositional 3D head avatars from any limited capture in a minute-level runtime.
    \item We introduce a feed-forward aggregated transformer that extracts and integrates features from arbitrary views, together with two Gaussian decoding branches that disentangle facial surfaces from hair strands to address their distinct geometric characteristics, jointly improving fidelity and efficiency.
    \item Experimental results demonstrate that our model, trained on large-scale 3D datasets, achieves state-of-the-art reconstruction quality for novel identities under sparse inputs, while enabling real-time rendering and supporting hair transfer and stylized face editing.
\end{itemize}

%% file: sec/2_related_works.tex
\section{Related Work}
\label{sec:related-works}

\subsection{Animatable Head Avatar Modeling}

In recent years, remarkable progress has been made in modeling animatable 3D head avatars from monocular or multi-view images and videos. Early work based on 3DMM~\cite{siggraph19993dmm, tog2017flame} optimized dynamic geometry~\cite{cvpr2019voca, tog2021deca, cvpr2022emoca} and neural textures~\cite{tog2018deep-video-portaits, cvpr2022nha-from-mono}, forming the foundation for portrait animation. With the success of NeRF~\cite{eccv2020nerf} in scene reconstruction, many studies~\cite{cvpr2021nerface, cvpr2021headnerf, cvpr2022rignerf, cvpr2023avatar-from-monocular, fg2023flame-in-nerf, cvpr2023insta} applied it to head reconstruction, achieving fine details but requiring dense, high-quality data and long rendering times unsuitable for animation. More recently, point-based representations, especially 3DGS~\cite{tog20233dgaussian}, have been used by several methods~\cite{cvpr2023pointavatar, cvpr2024flashavatar, cvpr2024gaussianavatars, eccv2024HeadGaS, siggragh2024monogaussianavatar, tvcg2025gaussianhead, cvpr2025taoavatar} for identity-specific head generation combined with 3DMM for fast control of pose and expression, though such single-subject optimization still demands extensive data.

Recent paradigms aim to improve efficiency by training generalized models on large-scale video or synthetic datasets. Encoder–decoder approaches~\cite{tog2022ava, iccv2023preface, eccv2024gphm, nips2024gagavatar, 3dv2025headgap, cvpr2025gasp} learn latent identity features but usually require finetuning or post-processing for unseen subjects. Diffusion-based methods~\cite{cvpr2025cap4d, cvpr2025gaf, corr2024facelift, cvpr2025perse} extend multi-view coverage through iterative optimization, achieving high quality but with heavy computational costs and possible 3D inconsistency. Transformer-based models~\cite{cvpr2024dust3r, corr2025vggt, iclr2024lrm, eccv2024lgm} have achieved impressive feed-forward 3D reconstruction of scenes and objects; however, their application to human faces and bodies, which require higher accuracy and controllability, remains limited. LAM~\cite{corr2025lam} and LHM~\cite{corr2025lhm} reconstruct faces and bodies from a single frontal image but show degraded side-view quality, while Avat3r~\cite{corr2025avat3r} predicts heads from four fixed views and synthesizes expressions via cross-attention layers, but its fixed input configuration restricts generality.

\input{figs/pipeline-overview}

\subsection{Strand-Based Hair Modeling}

Modeling hair separately remains a long-standing challenge due to its complex and highly variable geometry. Early work explored a wide range of representations, including 2D parametric surfaces~\cite{eg2000real-time-hair, pg2003hair-animation, cgi2004hair-nurbs}, explicit cylinders and strands~\cite{tog2004hair-multi, cvpr2019strand-mvs, eg2021hair-inverse}, as well as their combinations with neural volumetric fields~\cite{eccv2022neural-strands, iccv2023neural-haircut, cvpr2023neuwigs}. Other studies proposed purely implicit field representations~\cite{corr2023delta, tog2018hair-vae, 3dv2024teca} or mesh-based models~\cite{tog2009hair-meshes, tog20253dgh, corr2024hhavatar}. Several recent methods~\cite{corr2025hairgs, eccv2024gaussian-haircut, corr2024gaussian-hair} employ 3DGS as a geometric proxy for hair strands, producing highly detailed results but typically requiring an optimization stage. More recently, Perm~\cite{iclr2025perm} learns a parametric hair model from synthetic datasets, while DiffLocks~\cite{corr2025difflocks} trains a generative strand-level geometric prior on large-scale monocular data.

Unlike previous approaches, our model explicitly separates and independently models face and hair, while maintaining a unified framework within a composite transformer backbone and merged 3DGS representation. It requires neither dense or fixed capture setups, nor any time-consuming identity-specific training. Instead, with a single forward pass and an optional test-time refinement stage, our method reconstructs a high-quality face–hair composed 3DGS avatar from a few casual input images.

%% file: figs/pipeline-overview.tex
\begin{figure*}[t]
   \centering
   \includegraphics[width=1.0\linewidth]{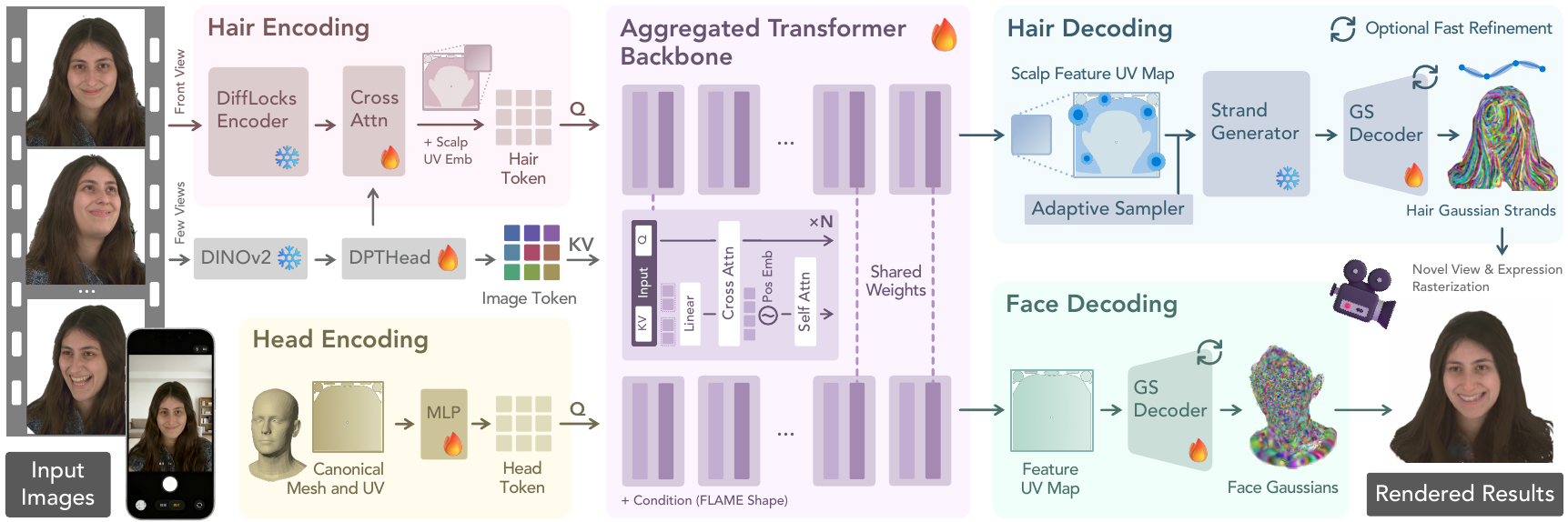}
   \vspace{-1.5em}
   \caption{\textbf{Pipeline Overview.} 
\paperName reconstructs a compositional face-and-hair 3D Gaussian head in the UV space. Our model starts with encoding image, hair, and face tokens from arbitrary input images and a template head mesh (\cref{sec:encoder}), which are fed into the aggregated transformer backbone to perform attention-based multi-view feature aggregation (\cref{sec:transformer}). The dual-branch decoders then independently decode planar Gaussians for the face and strand-based Gaussians for the hair at UV pixels, which are combined for real-time rendering under novel views and expressions (\cref{sec:dual-branch-decoder}).}
   \label{fig:pipeline-overview}
   \vspace{-0.5em}
\end{figure*}

%% file: sec/4_method.tex
\section{Method}
\label{sec:method}

Given a few images $\mathbf{I} = \{\mathbf{I}_1, \cdots, \mathbf{I}_{N}\}$, 
the goal is to rapidly predict a high-quality hair-compositional 3D head avatar $\mathcal{G}$ with explicitly separated hair and face Gaussian sets, which can be animated in real time to novel expressions and viewpoints. 
We further consider that the input size $N$ is arbitrary and unordered, as is typical in real-world user captures, posing the challenge of learning a flexible mapping from casual observations to a complete 3D head structure.

To overcome the challenge and achieve the desired goal, \paperName establishes a feed-forward pipeline as illustrated in \cref{fig:pipeline-overview}. It consists of three main components: efficient image and geometry encoders that extract multi-level facial and hairstyle features (\cref{sec:encoder}); a vision transformer-based backbone that aggregates multi-view features (\cref{sec:transformer}) and decodes them into planar face Gaussians and strand-based hair Gaussians (\cref{sec:dual-branch-decoder}); and training objectives that improve synthesis quality while enforcing clear region separation (\cref{sec:losses}). An optional refinement step can further enhance subject-specific fidelity in a minute-scale runtime.

\subsection{Model Architecture}

\subsubsection{Feature Tokenizer}
\label{sec:encoder}

The pipeline processes three types of tokens: (1) image tokens that directly extract texture and appearance details from the input images; (2) head geometric tokens that encode structural priors of the human head; and (3) hair tokens that capture geometric guidance for hair modeling. 

\paragraph{Image Feature Tokenization.}

To convert input images into transformer-compatible tokenized features, we utilize DINOv2~\cite{tmlr2024dinov2}, a vision foundation model pretrained on large-scale image datasets. A frozen DINOv2 backbone extracts multi-scale representations, which are further refined by a trainable and simplified DPTHead~\cite{iccv2021dpt} to produce image tokens $\mathbf{T}_{\text{image}}$ for the arbitrary $N$ input captures $\mathbf{I}_i$:
\begin{equation}
\mathbf{T}_{\text{image}} = \text{DPTHead} \left( \text{DINOv2} \left( \mathbf{I} \right)\right) \in \mathbb{R}^{N \times P \times C},
\end{equation}
where $P$ denotes the number of patch tokens per image and $C$ is the feature dimension. Through residual convolution and related operations applied to the multi-layer features within the DPTHead, the image tokens encode both high-frequency texture details from shallow layers and global appearance priors from deeper layers.

\paragraph{Head Geometric Prior Encoding.}

Given the head geometry prior provided by FLAME~\cite{tog2017flame}, we project the 3D spatial coordinates of the canonical template mesh vertices onto the UV space to obtain a position map via interpolation. For each pixel on the UV map, denoted as $\left\{ \mathbf{x}_i \right\}_{i=1}^{HW_{\text{uv}}} \subset \mathbb{R}^3$, we apply positional encoding $\gamma(\cdot)$ ~\cite{eccv2020nerf} followed by an MLP to produce the learnable head geometric tokens:
\begin{equation}
\mathbf{T}_{\text{head}} = \text{MLP} \left( \gamma(\mathbf{X}) \right) \in \mathbb{R}^{H_{\text{uv}} \times W_{\text{uv}} \times C}.
\end{equation}

\paragraph{Hair Feature Tokenization.}

We first select the image $\mathbf{I}_{\text{f}}$ that is closest to the frontal view and encode it using DiffLocks~\cite{corr2025difflocks}, a monocular hair prediction model pretrained on synthetic data, to extract the hair feature $f_{\text{hair}}$. Although DiffLocks can directly predict strand-based geometry, such monocular estimates often contain structural inaccuracies. To leverage multi-view consistency, we perform a cross-attention operation between $f_{\text{hair}}$ and the previously extracted image tokens, yielding the hair feature token $\mathbf{T}_{\text{hair}}$:
\begin{equation}
\begin{aligned}
f_{\text{hair}} &= \text{DiffLocks} \left( \mathbf{I}_{\text{f}} \right), \\
\mathbf{T}_{\text{hair}} &= \text{CrossAttn}\left( \mathbf{T}_{\text{image}}, f_{\text{hair}} \right) + \mathbf{T}_{\text{head}}^{\text{scalp}}.
\end{aligned}
\end{equation}
The resulting features are defined in the FLAME UV space over the scalp region, where $\mathbf{T}_{\text{head}}^{\text{scalp}}$ denotes the scalp subset of $\mathbf{T}_{\text{head}}$ and provides a spatial positional encoding, sharing the same feature dimension as $\mathbf{T}_{\text{hair}}$.

\subsubsection{Multi-View Fusion Transformer Backbone}
\label{sec:transformer}

At the heart of our reconstruction pipeline lies the aggregated transformer block, which jointly fuses image features with head and hair token representations to infer projection relationships from multi-view images, extract texture features, and compensate for structural deviations relative to the template head mesh. Within each transformer block, the head or hair tokens act as queries that attend to all image tokens via cross-attention. Inspired by VGGT~\cite{corr2025vggt}, we reshape the $N$ input images into the batch dimension to perform frame-wise self-attention. The forward pass for the head tokens is formulated as:
\begin{equation}
\begin{aligned}
\mathbf{T}_{\text{head}} &\leftarrow \text{CrossAttn}(\mathbf{T}_{\text{head}}, \mathbf{T}_{\text{image}} ;\boldsymbol{\psi}), \\
\mathbf{T}_{\text{image}_i} &\leftarrow \text{SelfAttn}(\mathbf{T}_{\text{image}_i}, \mathbf{T}_{\text{image}_i}),
\end{aligned}
\end{equation}
where $\boldsymbol{\psi}$ denotes the FLAME expression parameters tracked from the input images, concatenated with the image tokens along the feature dimension.

The same procedure applies to hair tokens $\mathbf{T}_{\text{hair}}$. The whole backbone network consists of multiple aggregated transformer blocks, where the cross-attention layers for $\mathbf{T}_{\text{head}}$ and $\mathbf{T}_{\text{hair}}$ are independent, while half of the self-attention layers are shared to promote feature coherence between head and hair representations.

\subsubsection{Dual-Branch Gaussian Decoder}
\label{sec:dual-branch-decoder}

\paragraph{Face Gaussian Branch.}

For the facial region, the transformer output feature $\mathbf{T}_{\text{head}}$ is decoded into Gaussian parameters. The decoders $\mathcal{D}$ for different parameters share the same convolutional backbone but have separate MLP heads for the specific attributes, including position offset $\Delta\mathbf{p} \in \mathbb{R}^3$, covariance $\boldsymbol{\sigma}$, rotation $\mathbf{r}$, opacity $\alpha$, and color $\mathbf{c}$:
\begin{equation}
\left\{ \Delta\mathbf{p}, \boldsymbol{\sigma}, \mathbf{r}, \alpha, \mathbf{c} \right\} = \mathcal{D}_{{\text{face}}\{\Delta\mathbf{p}|\boldsymbol{\sigma}|\mathbf{r}|\alpha|\mathbf{c}\}}(\mathbf{T}_{\text{head}}).
\end{equation}
Since the geometric tokens are defined in the FLAME UV space, this decoding process effectively predicts one Gaussian per UV-space pixel, which is then bound to the corresponding triangle face based on UV mapping. The offset $\Delta\mathbf{p} \in \mathbb{R}^3$ specifies the local displacement relative to the associated mesh triangle. To handle novel target expressions, the predicted Gaussians are transformed following the FLAME blendshape process. Adjusting the UV-map resolution allows control over the number of Gaussians, balancing visual quality and performance.

\paragraph{Hair Gaussian Branch.}

For the hair, which corresponds to the scalp region on the mesh, a single Gaussian per UV-space pixel cannot capture details such as curls or long hair. We therefore employ a dedicated branch to construct a strand of Gaussians for each pixel. Specifically, we add $\mathbf{T}_{\text{hair}}$ as a correction term to the original feature $f_{\text{hair}}$. The combined features for each UV pixel are decoded by the frozen strand generator $\mathcal{D}_{\text{dir}}$, composed of several modulated SIREN~\cite{iccv2021siren} layers from DiffLocks~\cite{corr2025difflocks}, into $S = 256$ direction vectors:
\begin{equation}
\label{eq:hair-dir-decode}
\mathbf{d}_{1:S} = \mathcal{D}_{\text{dir}}\left( \gamma \mathbf{T}_{\text{hair}} + f_{\text{hair}} \right),
\end{equation}
where $\gamma$ is a regularization coefficient for the correction term. Starting from the scalp vertex as the root, these directions are iteratively accumulated to form connected line segments, serving as a geometric prior for a strand of hair:
\begin{equation}
\mathbf{v}_s = \mathbf{v}_{s-1} + \mathbf{d}_s, \quad s=1,\dots,S.
\end{equation}
Each line segment in the strand is assigned a strand Gaussian, with its rotation and position aligned to the segment's direction and midpoint. The long axis of the scale equals half the segment length, and the two short axes are fixed to a small pre-defined radius $r$:
\begin{equation}
\begin{aligned}
\mathbf{p}_s &= \tfrac{1}{2}(\mathbf{v}_{s-1} + \mathbf{v}_s), \\
\boldsymbol{\sigma}_s &= \big[\,\tfrac{1}{2}\,\lVert \mathbf{d}_s \rVert_2,\ r,\ r\,\big]^{\top}.
\end{aligned}
\end{equation} 

Similar to the face branch, the color and opacity of each Gaussian are decoded from $\mathbf{T}_{\text{hair}}$ as:
\begin{equation}
\{\alpha, \mathbf{c}\} = \mathcal{D}_{\text{hair}\{\alpha|\mathbf{c}\}}(\mathbf{T}_{\text{hair}}).
\end{equation}

Considering that too many Gaussians would burden the renderer, we adaptively reduce both the number of strands and the number of Gaussians per strand $S$ for different hairstyles. A scalp UV-space density map, decoded from $\mathbf{T}_{\text{hair}}$, guides strand downsampling according to the average strand length (short, medium, or long). As shorter hairstyles are detected, the $S$ is reduced, and the Gaussian radius $r$ is increased to maintain scalp coverage.

\paragraph{Rendering.}

Finally, we merge the Gaussians from the face and hair branches, transform them to align with the target expression, and render the resulting 3D representation into the RGB image $\hat{\mathbf{I}}$ via differentiable rasterization under the specified camera parameters.

\subsection{Loss Function and Fast Refinement}
\label{sec:losses}

The training objective integrates a hair region separation loss tailored to the dual-branch design, a full-image photometric reconstruction loss, and regularization penalties that constrain the geometric stability of Gaussian primitives. The overall loss function is formulated as:
\begin{equation}
\mathcal{L}_{\text{total}} = \mathcal{L}_{\text{hair}} + \mathcal{L}_{\text{photo}} + \mathcal{L}_{\text{reg}}.
\end{equation}

\paragraph{Hair Region Loss.}



To encourage a clear separation between the face and hair Gaussians, we adopt the parsing mask from~\cite{eccv2024sapiens} to extract the hair region $\mathbf{I}_{\text{hair}}$ and supervise the rendered image $\hat{\mathbf{I}}_{\text{hair}}$ generated only from the hair Gaussians via an L2 reconstruction loss. Additionally, following~\cite{mm2024gaussiantalker}, we perform semantic rendering on the dual-branch Gaussians to produce $\mathbf{I}_{\text{seg}}$ and impose an auxiliary L2 constraint, effectively mitigating the tendency of face Gaussians to occupy the top hair area in short-hair cases, thereby preserving strand-level geometric and visual consistency. The hair region loss is formulated as 
\begin{equation}
\mathcal{L}_{\text{hair}} = \lambda_{\text{hair}} \left\| \hat{\mathbf{I}}_{\text{hair}} - \mathbf{I}_{\text{hair}} \right\|_2 + \lambda_{\text{seg}} \left\| \hat{\mathbf{I}}_{\text{seg}} - \mathbf{I}_{\text{seg}} \right\|_2.
\end{equation}

\paragraph{Photometric Loss.}

We supervise the rendered RGB results under novel views or expressions using the corresponding ground-truth images. The photometric loss combines an L1 reconstruction loss, a structural similarity (SSIM) loss~\cite{tog20233dgaussian}, and a perceptual loss (LPIPS)~\cite{cvpr2018perceploss} to preserve high-frequency details:
\begin{equation}
\mathcal{L}_{\text{photo}} = \mathcal{L}_{1} + \lambda_{\text{ssim}} \mathcal{L}_{\text{ssim}} + \lambda_{\text{lpips}} \mathcal{L}_{\text{lpips}}.
\end{equation}

\paragraph{Regularization Term.}

The predicted position offsets and scaling factors for each Gaussian primitive must be carefully constrained; otherwise, unconstrained motion or expansion can introduce visual artifacts in both reconstruction and animation. We therefore define a regularization term that $\mathcal{L}_{\text{reg}}$ penalizes extreme values by applying a thresholded L2 penalty separately to position and scale as:
\begin{align}
\begin{aligned}
\mathcal{L}_{\text{reg}} = &\lambda_{\text{pos}} \left\| \max \left( 0, \bar{\mathbf{p}} - \epsilon_{\text{pos}} \right) \right\|_2^2 \\
&+ \lambda_{\text{scale}} \left\| \max \left( 0, \bar{\boldsymbol{\sigma}} - \epsilon_{\text{scale}} \right) \right\|_2^2.
\end{aligned}
\end{align}

\paragraph{Optional Fast Refinement.}

After learning priors from large-scale data, our pipeline can extract features of unseen identities from arbitrary inputs and produce a complete 3DGS model in a single forward pass. To further enhance person-specific details, we provide an optional fast refinement stage, where the encoder and transformer backbone are frozen, while the predicted aggregated tokens $\mathbf{T}_{\text{head}}$ and $\mathbf{T}_{\text{hair}}$, along with the dual-branch decoders, are jointly optimized. This lightweight refinement process, typically completed within a few minutes, can effectively improve reconstruction quality and better adapt to diverse identities and lighting conditions in the wild.

%% file: sec/5_experiments.tex
\section{Experiments}
\label{sec:experiments}

\input{tables/qualitative-comparison}
\input{figs/qualitative-comparison}

\subsection{Experimental Settings}

\paragraph{Datasets.}

We take the NeRSemble~\cite{tog2023nersemble} dataset, which has been widely used in recent multi-view head reconstruction studies~\cite{3dv2025headgap, cvpr2025gaf, corr2025avat3r, cvpr2025gasp, cvpr2025mega, cvpr2024gaussianavatars, cvpr2025cap4d}. The dataset contains about 70k frames from 202 persons after filtering, captured by a 16-view camera system and covering diverse head poses and expression sequences. We adopt BackgroundMattingV2~\cite{cvpr2021bgmatting}, STAR~\cite{cvpr2023star}, and VHAP~\cite{cvpr2024gaussianavatars} to remove backgrounds, extract facial regions, crop out the subjects, resize them to $512 \times 512$, and estimate FLAME coefficients and monocular camera poses. We train on 195 identities and hold out 7 for testing.
For each training sample, we randomly select between 1 and 6 images as input, and use 4 images with different views and expressions for supervision, enabling our model to handle real-world scenarios with sparse and varying numbers of input frames. 
For evaluation, we use all remaining images except the inputs and report the average performance.

We also construct an in-house dataset captured with mobile phones, containing about 4k frames from 6 persons, as a supplementary test set to further evaluate the model’s performance on in-the-wild inputs.

\paragraph{Baselines.}

We evaluate \paperName under both single-view and multi-view reconstruction settings, comparing it against recent state-of-the-art 3D head avatar generation methods across three paradigms: 
(1) \textit{optimization-based} methods: GaussianAvatars~\cite{cvpr2024gaussianavatars}, FlashAvatar~\cite{cvpr2024flashavatar}, and MeGA~\cite{cvpr2025mega}; 
(2) \textit{feed-forward} methods: GAGAvatar~\cite{nips2024gagavatar} and LAM~\cite{corr2025lam}; 
(3) \textit{diffusion-based} methods: DiffusionRig~\cite{cvpr2023diffusionrig}.

\paragraph{Evaluation Metrics.}

We evaluate reconstructed Gaussian head avatars by focusing on reenactment performance under extensive novel views and expressions. For self-reenactment, where ground truth images are available, we quantitatively assess the realism and quality of the rendered results using three paired-image metrics: Peak Signal-to-Noise Ratio (PSNR), Structural Similarity Index (SSIM)~\cite{tip2004ssim}, and Learned Perceptual Image Patch Similarity (LPIPS)~\cite{cvpr2018lpips}. Additionally, we employ two face-specific metrics: the Average Keypoint Distance (AKD), computed from facial landmarks detected by PIPNet~\cite{ijcv2021pipnet}, and the cosine similarity of identity embeddings (CSIM) extracted using ArcFace~\cite{cvpr2019arcface}.

\paragraph{Implementation Details.}

We implement \paperName in PyTorch. The reconstruction transformer consists of 4 layers of aggregated blocks with 8 self-attention heads, 16 cross-attention heads, and a feature dimension of 1024. During training, we use the Adam optimizer~\cite{iclr2015adam} and a cosine learning rate scheduler with 600 warm-up iterations. The initial learning rate is set to $\text{1e\text{-}4}$. The loss weights are set as $\lambda_{\text{hair}} = \lambda_{\text{seg}} = 0.3$, $\lambda_{\text{ssim}} = 0.5$, $\lambda_{\text{lpips}} = 0.02$, and $\lambda_{\text{pos}} = \lambda_{\text{scale}} = 0.1$. We train with a batch size of 1 on two NVIDIA H20 GPUs for 50,000 iterations, which takes approximately 4 days.

For the fast refinement stage, we fine-tune our Gaussian decoder for 100 epochs on the input data. For evaluation, we use the official and publicly available implementations of all baselines with their recommended settings. All these experiments are conducted on one single consumer-grade NVIDIA RTX 4090 GPU (24\,GB).

\subsection{Main Results}


\cref{table:qualitative-comparison} presents both single-shot and few-shot reconstruction results on the NeRSemble dataset. Compared with recent state-of-the-art models, our method achieves the best image reconstruction quality across all input settings, as indicated by the PSNR, SSIM, and LPIPS metrics, with PSNR improvements of 1.72, 0.27, and 0.72 under 3, 6, and 16-frame inputs, respectively.
Moreover, our approach accurately predicts novel expressions and poses, as revealed by the AKD metric, achieving as low as 3.66 for single frame and 2.72 for multiple frames input, demonstrating strong reenactment capability.  
Additionally, our model maintains the highest identity consistency across multiple viewpoints, indicating that the learned priors can successfully predict complete head geometry even under sparse observations—consistent with the qualitative comparisons shown in \cref{fig:qualitative-comparison}.

\cref{fig:qualitative-comparison} further illustrates that, on both the NeRSemble and our in-house phone-captured in-the-wild datasets, \paperName effectively reconstructs the geometric and textural characteristics of unseen identities, demonstrating robust adaptability to diverse real-world capture scenarios.  
Notably, our model uniquely establishes a strand-connected, independent set of hair Gaussians that aligns with the physical structure of real hair.  
Meanwhile, existing optimization-based methods such as GaussianAvatars, FlashAvatar, and MeGA often fail to form a complete 3D head under sparse inputs, even collapsing at viewpoints or expressions similar to those in the inputs.  
DiffusionRig, as a 2D diffusion-based method, can synthesize visually similar faces but produces low-resolution renderings and frequently violates camera constraints, leading to positional misalignment—explaining its significantly higher AKD.  
Previous feed-forward approaches such as LAM and GAGAvatar can generate animatable 3D avatars from a single image, but they struggle to maintain multi-view identity consistency and accurately control novel head poses, which is reflected by their relatively competitive SSIM yet worse AKD and PSNR.

In terms of runtime efficiency,  our approach achieves minute-level multi-view reconstruction, running 10×–100× faster than optimization-based methods and reaching an excellent balance between quality and speed.
While single-view approaches such as LAM and GAGAvatar achieve instant modeling, their visual quality degrades.
However, our model already produces leading reconstruction performance with a single forward pass (w/o refinement) in a comparable runtime.
Moreover, by decoupling coordinate computation from the existing Gaussian-splatting renderer, our avatars support real-time animation at up to 250 FPS.

\input{tables/ablation-study}
\input{figs/ablation-study}

\subsection{Ablation Study}

\paragraph{Effect of Strand-Based Hair Branch.}

To verify the effectiveness of our dual-branch design, we remove the hair branch and revert the model to predicting the entire Gaussian head from the FLAME UV map.  
As shown in \cref{table:ablation-study}, this results in a performance degradation of 1.82, 0.014, and 0.088 in PSNR, SSIM, and LPIPS, respectively.  
As illustrated in \cref{fig:ablation-study}, Gaussians anchored to the FLAME template cannot sufficiently represent long-hair regions, and the feed-forward model fails to adapt the number of Gaussians, producing blurred, non-strand-like hair geometry.  
These results confirm that modeling flexible hair and fixed facial regions separately is a natural and effective design choice.

\paragraph{Effect of Hair Region Loss.}

\cref{fig:ablation-study} (right) shows that without the hair region loss supervision, the hair, especially near the top, fails to clearly separate from the scalp, degrading both reconstruction quality and downstream applications such as hairstyle transfer.

\paragraph{Effect of Fast Refinement.}

As shown in \cref{table:ablation-study}, the additional fast refinement stage leads to improvements of 1.29, 0.016, and 0.049 in PSNR, SSIM, and LPIPS metrics, respectively.  
\cref{fig:ablation-study} demonstrates how the refinement further extracts and restores more details of the input subject, building upon the complete, subject-specific avatar already generated in a single forward pass of our model (w/o finetune).

\paragraph{Incremental Reconstruction.}

\paperName supports input sequences with varying numbers of images, viewpoints, and expressions, offering greater generality and flexibility compared to previous methods that require fixed input numbers or camera views. As shown in \cref{fig:incremental-reconstruction}, reconstruction quality increases rapidly as the number of input frames grows from 1 to 6, and gradually plateaus beyond that point. Qualitatively, additional inputs allow the model to capture finer dynamic details such as the eye sockets and teeth. Combined with our decoupled fast refinement and real-time rendering process, this enables progressive enhancement of reconstruction quality.

\input{figs/incremental-reconstruction}

\subsection{More Applications}

\paragraph{Hairstyle Transferring.}

As shown in \cref{fig:hairstyle-transferring}, our method allows for the simple alteration of A's hairstyle with B's. Specifically, we load A's face branch Gaussian and B's hair branch Gaussian, both of which correspond to the unified scalp UV space. By applying the positional offset between corresponding vertices on A and B's shaped FLAME, we make small adjustments to the strand roots, aligning B's hair with A's without extra optimization.

\input{figs/hairstyle-transferring}

\paragraph{Convenient Face Stylize Editing.}

\cref{fig:stylize-editing} shows the impressive results of applying our framework to stylized editing on the face. Unlike previous approaches~\cite{corr2025lam, cvpr2025mega} that require preprocessing inputs, re-inference, or iterative training for stylization, \paperName binds the Gaussians to the UV space, meaning that editing the 2D texture maps can directly lift up to the 3D Gaussian space, enabling convenient customization of animatable avatars with full 3D consistency.

\input{figs/stylize-editing}

%% file: tables/qualitative-comparison.tex
\newcommand{\goldcell}{\cellcolor[HTML]{F3E598}}
\newcommand{\silvercell}{\cellcolor[HTML]{D7D7D7}}

\begin{table*}[h]
    \caption{{\bf Quantitative Results} under both single-view and multi-view input settings. Note that our model was \textbf{trained once} on mixed input numbers to learn generalizable priors. Modeling time denotes the time required to reconstruct the 3DGS model, excluding the estimation of driving parameters that can be precomputed in advance, while FPS corresponds to the frame rate during the animation rendering process of the output model. All methods are rendered on a black background for metrics calculation.}
    \vspace{-0.4em}
    \centering
    \renewcommand{\arraystretch}{0.9}
    \scalebox{1}{
        \begin{tabular}{cl|ccccc|c|c}
        \toprule
        & Method & PSNR$\uparrow$ & SSIM$\uparrow$ & LPIPS$\downarrow$ & AKD$\downarrow$ & CSIM$\uparrow$ & Modeling Time$\downarrow$ & FPS$\uparrow$ \\ \midrule
        \multirow{5}{*}{1 frame} 
        & DiffusionRig~\cite{cvpr2023diffusionrig}
                 & 16.39& 0.256&  0.487&56.48&0.423& $\sim$ 7\,m& 0.85   \\
        & GAGAvatar~\cite{nips2024gagavatar}
                 & 14.19& 0.568&  0.522&84.10&\underline{0.599}& 1.54\,s& 44.16\\
        & LAM~\cite{corr2025lam}
                 & 16.41& 0.662&  0.409&48.64&0.461& \textbf{0.31\,s}& 142.77\\
        & Ours~(single-pass) & \underline{22.46} & \textbf{0.803}&  \underline{0.325} & \underline{3.96} & 0.522& 3.13\,s& 258.36\\ 
        & Ours~(full) & \textbf{22.80}& \underline{0.797} &  \textbf{0.303}& \textbf{3.66}& \textbf{0.665}& 49.01\,s& \textbf{259.56}\\ \midrule
        \multirow{5}{*}{3 frames} 
        & DiffusionRig~\cite{cvpr2023diffusionrig}
                 & 16.29& 0.271&  0.492&56.59&\underline{0.524}& $\sim$ 14\,m & 0.89   \\
        & FlashAvatar~\cite{cvpr2024flashavatar}
                 & 18.10& 0.659&  0.364&51.07&0.161& $\sim$ 1\,h& 220.45\\
        & GaussianAvatars~\cite{cvpr2024gaussianavatars}
                 & \underline{21.68}& \underline{0.747}&  \underline{0.328}&\underline{13.00}&0.347& $\sim$ 1.2\,h& 47.23\\
        & MeGA~\cite{cvpr2025mega}
                 & 16.63& 0.607&  0.459&29.66&0.236& $\sim$ 4\,h& 33.64\\
        & Ours~(full) & \textbf{23.40}& \textbf{0.813}&  \textbf{0.295}& \textbf{3.37}&\textbf{0.700}& \textbf{49.23\,s}& \textbf{258.22}\\ \midrule
        \multirow{5}{*}{6 frames} 
        & DiffusionRig~\cite{cvpr2023diffusionrig}
                 & 16.55& 0.276&  0.490&56.27&\underline{0.586}& $\sim$ 18\,m& 0.41    \\
        & FlashAvatar~\cite{cvpr2024flashavatar}
                 & 16.08& 0.601&  0.437&72.91&0.074& $\sim$ 1.5\,h & 217.55\\
        & GaussianAvatars~\cite{cvpr2024gaussianavatars}
                 & \underline{23.44}& \underline{0.784}&  \underline{0.300}& \underline{7.41}&0.465& $\sim$ 1.5\,h & 48.15\\
        & MeGA~\cite{cvpr2025mega}
                 & 17.29& 0.680&  0.346&13.30&0.290& $\sim$ 5\,h& 34.98\\
        & Ours~(full) & \textbf{23.71}& \textbf{0.825}&  \textbf{0.296}& \textbf{3.08}&\textbf{0.721}& \textbf{52.15\,s}& \textbf{246.92}   \\ \midrule
        \multirow{5}{*}{16 frames} 
        & DiffusionRig~\cite{cvpr2023diffusionrig}
                 & 16.21& 0.257&  0.528&56.70&\underline{0.673}& $\sim$ 36\,m& 0.49    \\
        & FlashAvatar~\cite{cvpr2024flashavatar}
                 & 15.84& 0.612&  0.447&70.06&0.055& $\sim$ 1.5\,h & 198.12 \\
        & GaussianAvatars~\cite{cvpr2024gaussianavatars}
                 & \underline{23.50}& \underline{0.769}&  \textbf{0.152}& \underline{5.54}&0.629& $\sim$ 4.2\,h& 32.79 \\
        & MeGA~\cite{cvpr2025mega}
                 & 17.98& 0.729&  0.290& 10.00& 0.446& $\sim$ 5\,h& 28.43 \\
        & Ours~(full) & \textbf{24.22}& \textbf{0.837}& \underline{0.271}& \textbf{2.72}&\textbf{0.770}& \textbf{130.12\,s}& \textbf{242.21}    \\
        \bottomrule
        \end{tabular}}
    \label{table:qualitative-comparison}
\end{table*}

%% file: figs/qualitative-comparison.tex
\begin{figure*}[t]
   \centering
   \includegraphics[width=1.0\linewidth]{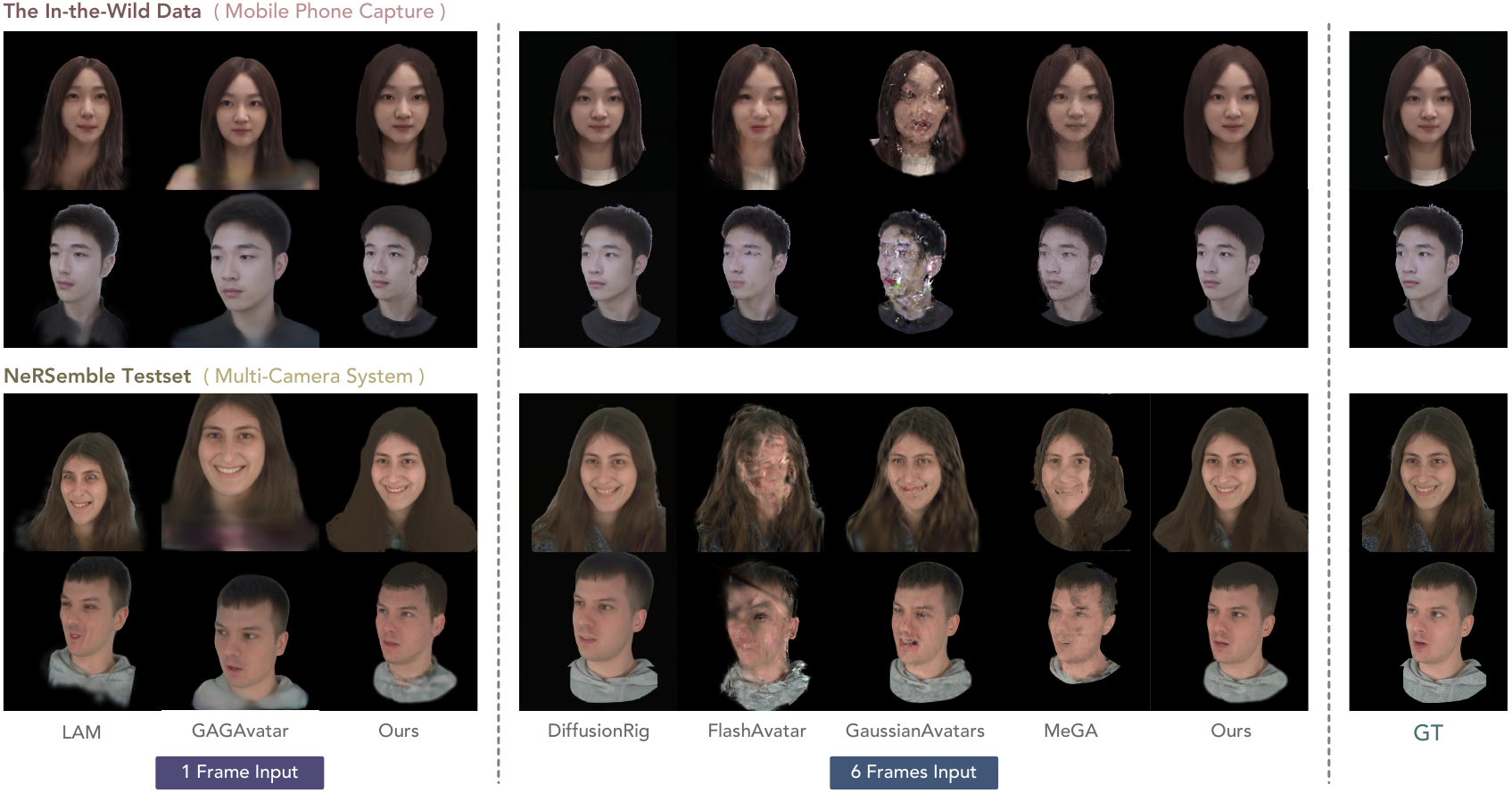}
   \vspace{-1.7em}
   \caption{\textbf{Qualitative Comparison} on reconstructing unseen identities from both in-the-wild data and the NeRSemble dataset under different capture conditions. LAM struggles to preserve identity similarity, while GAGAvatar and DiffusionRig fail to maintain accurate control under novel expressions or viewpoints. Optimization-based methods such as GaussianAvatars, FlashAvatar, and MeGA often fail to fit under sparse inputs. In contrast, our method delivers high rendering quality, supports accurate expression reenactment, and maintains consistent identity.}
   \vspace{-1.0em}
   \label{fig:qualitative-comparison}
\end{figure*}

%% file: tables/ablation-study.tex
\begin{table}[t]
    \caption{{\bf Ablation Study}.}
    \vspace{-0.5em}
    \centering
    \scalebox{1.0}{
        \begin{tabular}{l|cccc}
        \toprule
        Method & PSNR$\uparrow$ & SSIM$\uparrow$ & LPIPS$\downarrow$ \\ \midrule
        w/o Hair Branch & 23.32& 0.782& 0.421\\
        w/o Region Loss & 25.04& 0.794& 0.336\\
        w/o Finetune    &  23.85& 0.780& 0.382\\ 
        Full            & \textbf{25.14}& \textbf{0.796}& \textbf{0.333}\\ 
        \bottomrule
        \end{tabular}}
    \label{table:ablation-study}
\end{table}

%% file: figs/ablation-study.tex
\begin{figure}[t]
   \centering
   \includegraphics[width=1\linewidth]{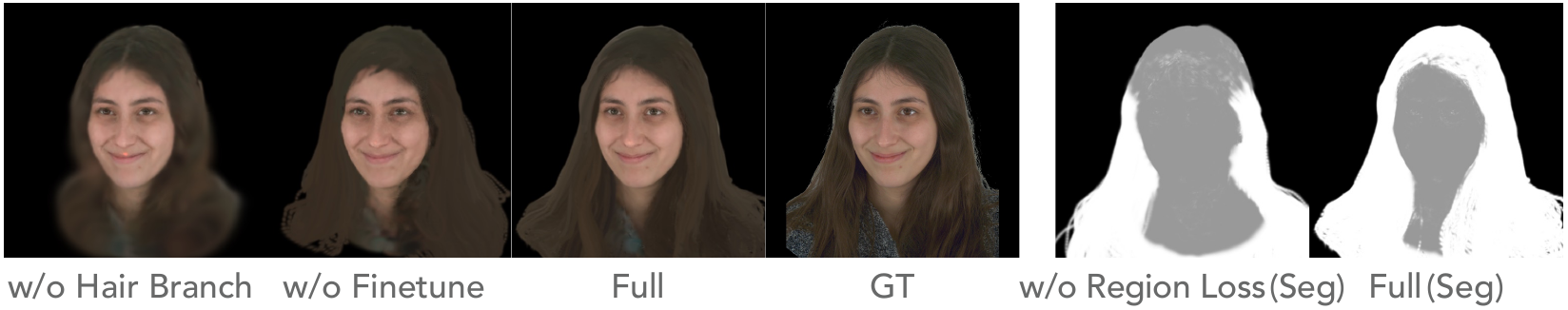}
   \vspace{-1.5em}
   \caption{\textbf{Ablation Study.} Best viewed with zoom-in.}
   \vspace{-1.5em}
   \label{fig:ablation-study}
\end{figure}

%% file: figs/incremental-reconstruction.tex
\begin{figure}[h]
    \vspace{-0.5em}
   \centering
   \includegraphics[width=0.9\linewidth]{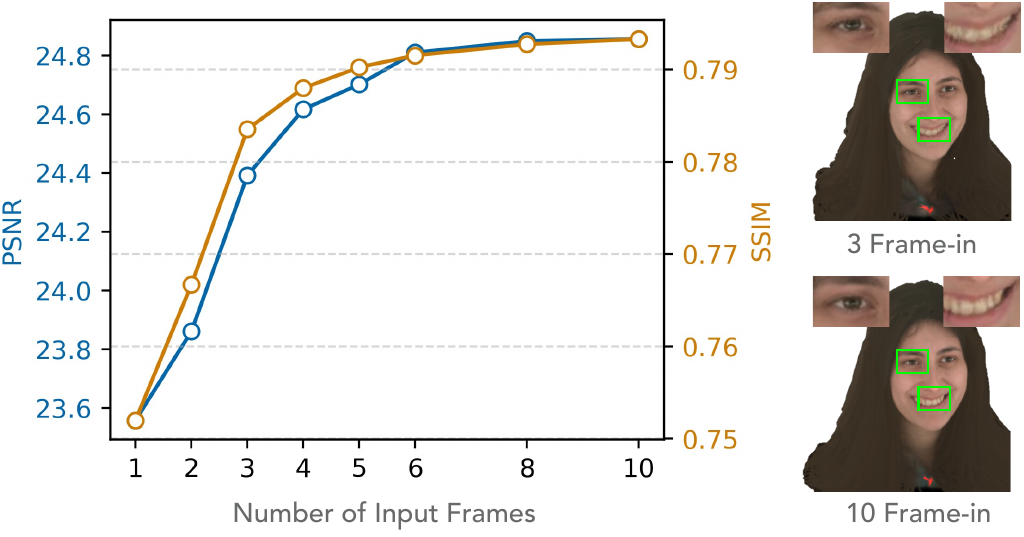}
   \vspace{-0.5em}
   \caption{Reconstruction quality improves with additional frames, helping the model capture finer details. Note that for varying input quantities, we perform refinement over the same epochs.}
   \label{fig:incremental-reconstruction}
   \vspace{-0.5em}
\end{figure}

%% file: figs/hairstyle-transferring.tex
\begin{figure}[t]
   \centering
   \includegraphics[width=0.9\linewidth]{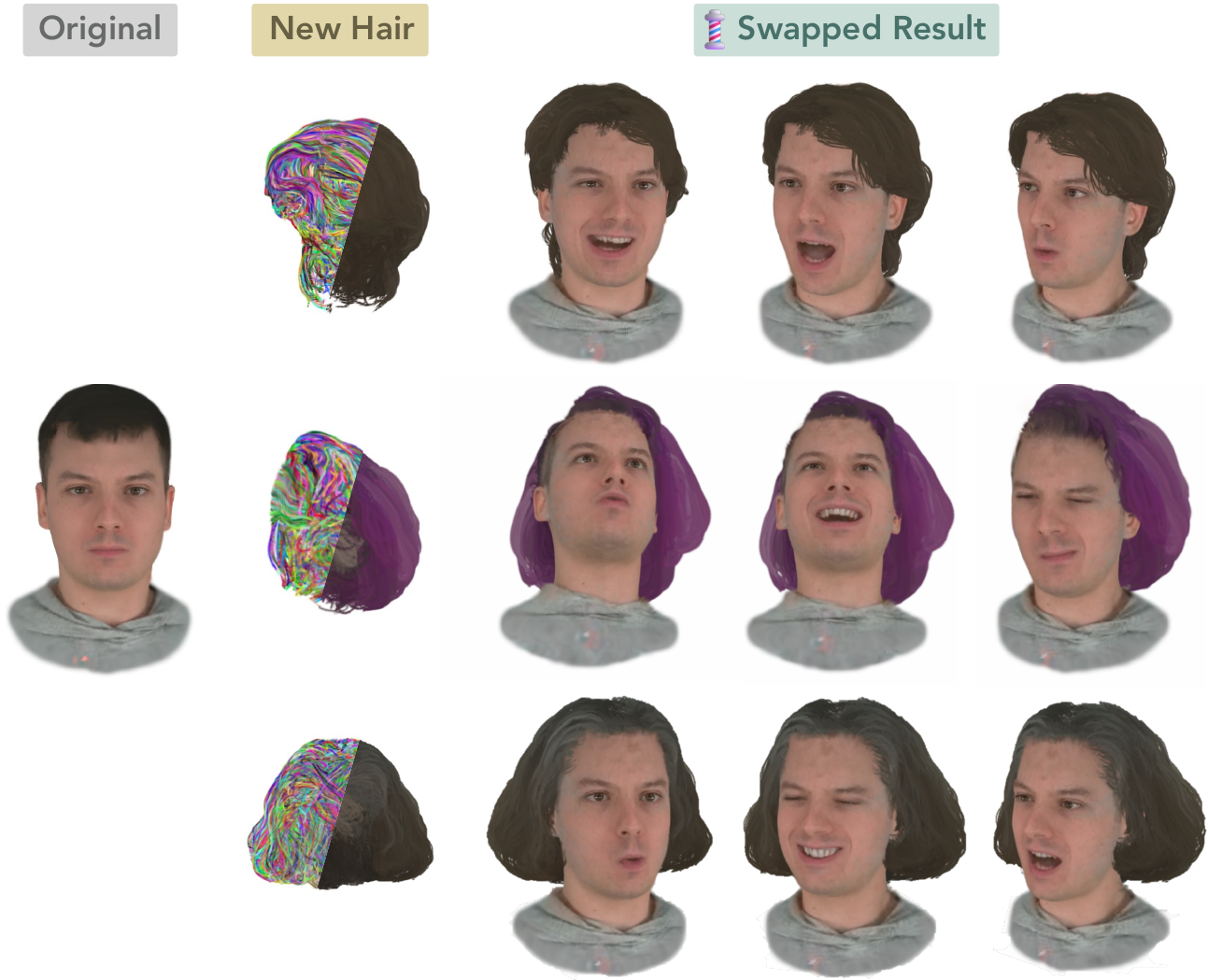}
   \vspace{-0.3em}
   \caption{\textbf{Hairstyle Transferring.} The compositional face-and-hair dual-branch design enables seamless hairstyle transfer between our avatars, even across different genders.}
   \vspace{-1.0em}
   \label{fig:hairstyle-transferring}
\end{figure}

%% file: figs/stylize-editing.tex
\begin{figure}[t]
   \centering
   \includegraphics[width=0.9\linewidth]{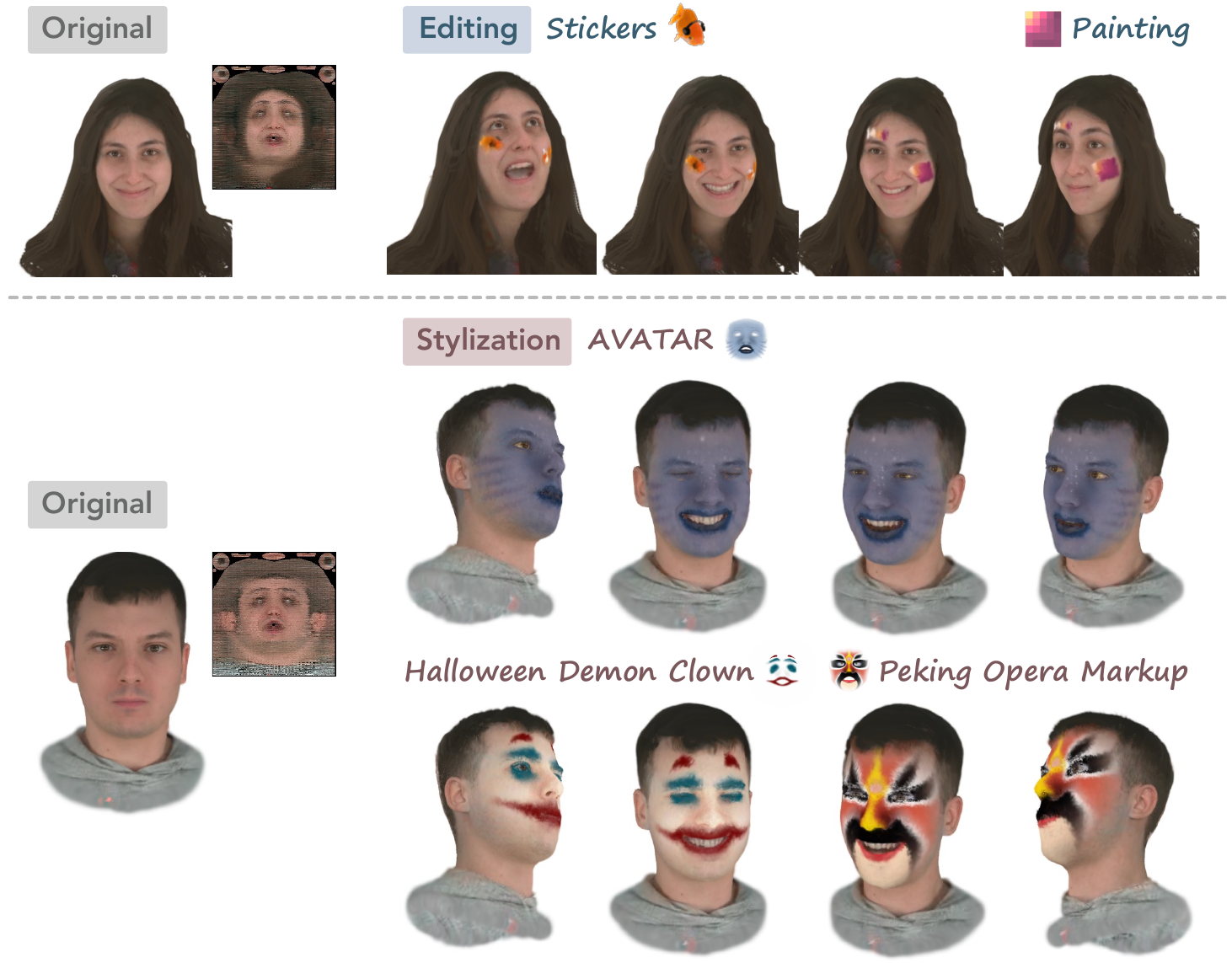}
   \vspace{-0.3em}
   \caption{\textbf{Stylize Editing.} Texture-based editing or stylization on the reconstruction result enables convenient 3D-aware manipulation without re-inference.}
   \label{fig:stylize-editing}
   \vspace{-1.5em}
\end{figure}

%% file: sec/6_conclusion.tex
\section{Conclusion}
\label{sec:conclusion}

In this work, we present \paperNameWOSpace{}, a novel model for fast and high-quality 3DGS hair and face generation. We demonstrate that, with priors learned from large-scale 3D head data, two key designs—an aggregated transformer and a dual-branch Gaussian decoder for face and hair—enable the model to capture fine-grained geometric and texture details across varying numbers of input views. Through both quantitative and qualitative evaluations, we show that \paperNameWOSpace{} can reconstruct unseen identities within minutes and achieve real-time animation under novel poses and expressions, without the need for traditional studio-level data acquisition or long optimization. Moreover, thanks to the dual-branch design and UV-space binding, our model supports user-friendly post-applications such as hairstyle transferring and texture editing.

%% file: sec/X_suppl.tex
\clearpage
\setcounter{page}{1}
\maketitlesupplementary

\begin{appendix}

\section{Implementation Details}
\label{sec: sup-impl-details}

In this section, we provide further details on the data pre-processing, model architecture, and experimental settings to facilitate reproducibility.

\subsection{Data Pre-Processing}

For data with varying capture conditions, our data pre-processing pipeline consists of four steps: (1) background removal, using the faster BackgroundMattingV2~\cite{cvpr2021bgmatting} on multi-view datasets with clear boundaries such as NeRSemble~\cite{tog2023nersemble}, and the more robust Sapiens~\cite{eccv2024sapiens} on user-captured monocular images and videos; (2) keypoint detection, using STAR~\cite{cvpr2023star} to obtain facial landmarks and locate the face; (3) FLAME~\cite{tog2017flame} parameters tracking following VHAP~\cite{cvpr2024gaussianavatars}; (4) based on tracking results, mesh projection and further cropping of the facial region to remove shoulders and torso.

To better match the diverse real-world data captured by smartphones, we improve the original VHAP with three modifications: First, we discard the global FLAME head rotation and only optimize the neck rotation, which avoids entanglement between global and local pose and makes the recovered head motion easier to reuse with full-body models such as SMPL. Second, we adopt a motion-aware iteration schedule: the number of optimization steps per iteration is increased when the average distance between the current landmarks and the reference frontal frame landmarks (determined by the distance between the eye landmarks) is large, and kept close to a small base value for iterations with little movement. Concretely, given the average distance $d_{\text{lmk}}$ between the landmarks relative to the reference frontal frame, we compute a per-iteration proposal as follows:
\begin{equation}
  N^{\text{cur}} =
  \begin{cases}
    N_0 + \left\lfloor \Delta \,(d_{\text{lmk}} - d_{\text{th}}) \right\rfloor, & d_{\text{lmk}} > d_{\text{th}}, \\
    N_0, & \text{otherwise},
  \end{cases}
\end{equation}
where $d_{\text{th}}$ is a distance threshold, $N_0$ is the base number of iterations, and $\Delta$ is a scaling factor. For monocular video, we then apply exponential smoothing with an upper bound:
\begin{equation}
  N_t = \min\left(
    \left\lfloor \lambda N_{t-1} + (1 - \lambda) N_t^{\text{cur}} \right\rfloor,
    N_{\max}
  \right),
\end{equation}
where $\lambda$ is the smoothing weight, and $N_{\max}$ is the maximum iteration budget. This allows the tracker to better follow rapid head movements while keeping the overall computational cost moderate.

\begin{figure}[t]
    \centering
    \begin{subfigure}{0.9\linewidth}
        \centering
        \includegraphics[width=\linewidth]{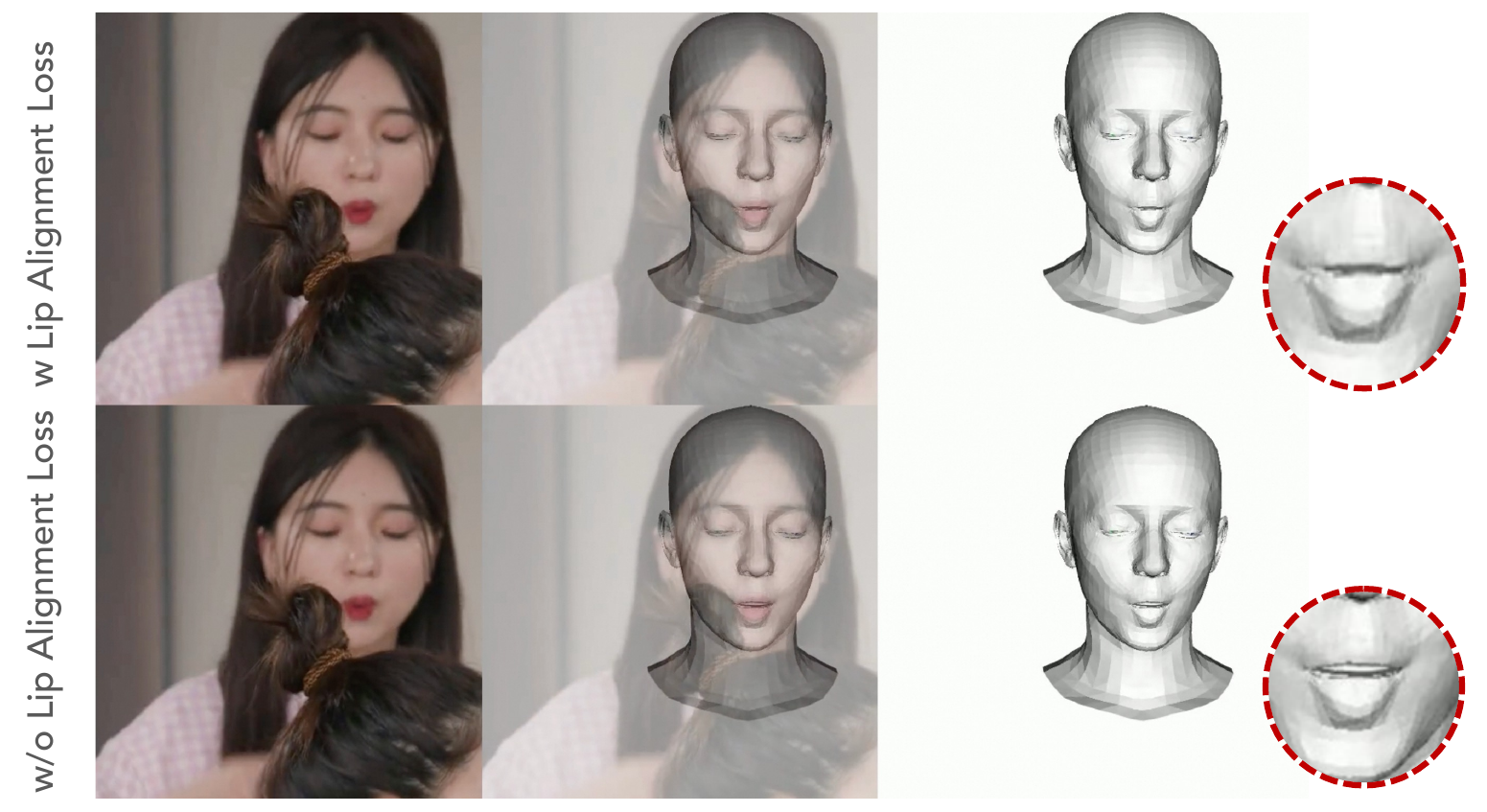}
        \caption*{(a)\quad Effect of lip alignment loss.}
    \end{subfigure}
    \vspace{-0.5em}
    \begin{subfigure}{0.9\linewidth}
        \centering
        \includegraphics[width=\linewidth]{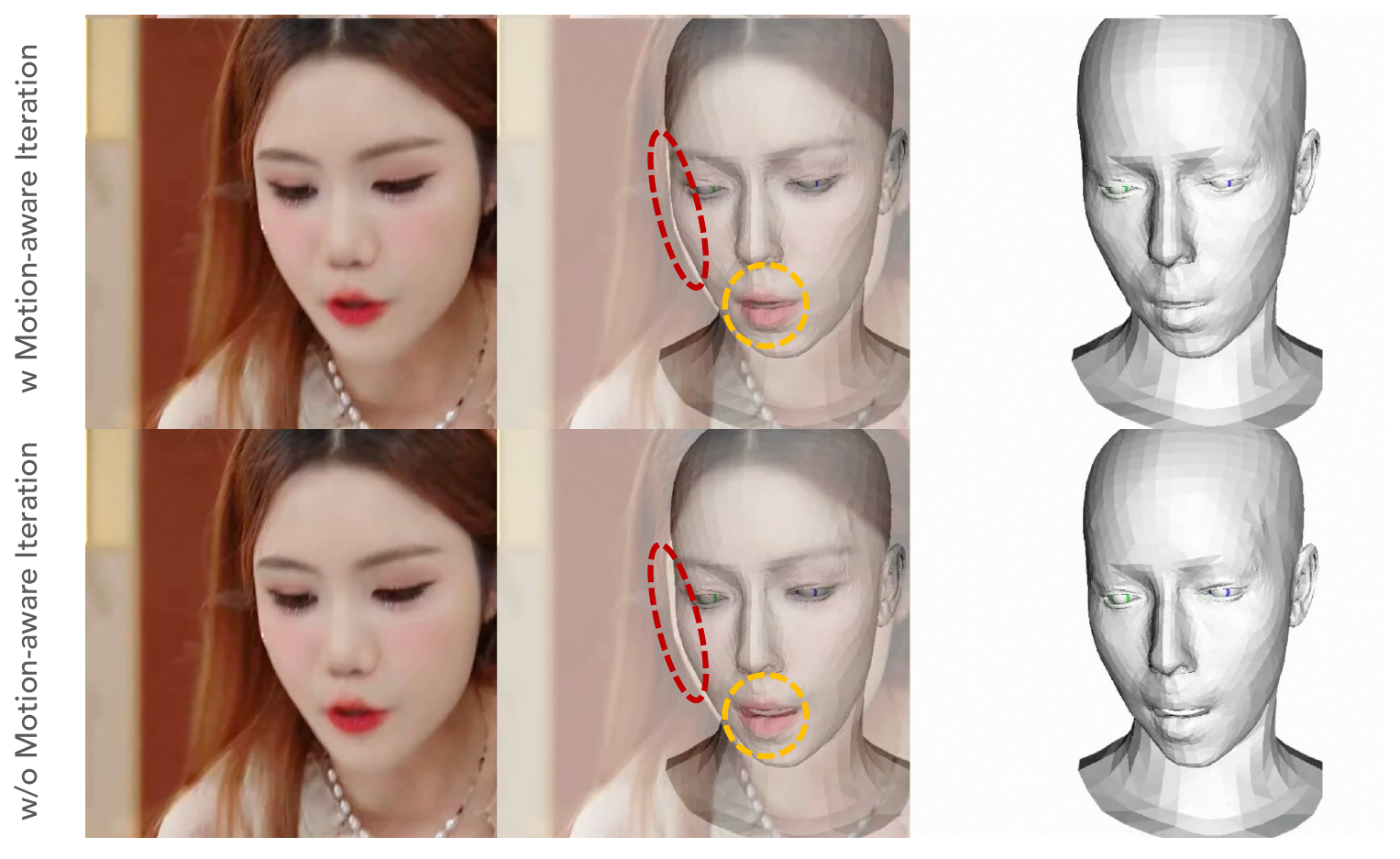}
        \caption*{(b)\quad Effect of motion-aware iteration schedule.}
    \end{subfigure}

    \vspace{-0.3em}
    \caption{\textbf{Effects of our preprocessing improvements}. These modifications make the lip movement and facial shape tracking more accurate.}
    \label{fig:preprocess-ablation}
    \vspace{-1.0em}
\end{figure}

Third, for the authenticity of novel expression reenactments, details such as eye closure and lip alignment are crucial. We introduce two additional structural perceptual loss terms: the eye closure loss $\mathcal{L}_{\text{eye}}$ and the lip alignment loss $\mathcal{L}_{\text{lip}}$. These losses not only enhance the supervision of landmark position errors but also penalize the discrepancy in distances between corresponding eye and lip landmarks.
\begin{align}
\begin{aligned}
\mathcal{L}_{\text{eye}} &= 
\gamma \cdot \left\| \hat{\mathbf{K}}_{\text{eye}} - \mathbf{K}_{\text{eye}}\right\|_2
+ (1 - \gamma) \cdot \left| \hat{d}_{\text{eye}} - d_{\text{eye}} \right|,
\end{aligned}
\end{align}

\begin{align}
\begin{aligned}
\mathcal{L}_{\text{lip}} &=
\gamma \cdot \left\| \hat{\mathbf{K}}_{\text{lip}} - \mathbf{K}_{\text{lip}} \right\|_2
+ (1 - \gamma) \cdot \left| \hat{d}_{\text{lip}} - d_{\text{lip}} \right|,
\end{aligned}
\end{align}
where $\gamma = 0.95$.  
$\mathbf{K}_{\text{lip}}$, $\hat{\mathbf{K}}_{\text{lip}}$, $\mathbf{K}_{\text{eye}}$, and $\hat{\mathbf{K}}_{\text{eye}}$ represent the ground truth and predicted landmark vectors for the lip and eye regions in both the original image and the rendered FLAME model.  
$d_{\text{eye}}$ and $d_{\text{lip}}$ correspond to the distances between the left and right eyes and the upper and lower lips, respectively.

\subsection{Network Architecture}

\paragraph{Adaptive Hair Sampler.}

In generating strand Gaussians for hair, we introduce an adaptive sampler strategy that uses a varying number of Gaussians for different hairstyles, reducing rendering pressure while maintaining quality. This optimization is achieved by downsampling both the total number of strands and the number of Gaussians per strand simultaneously: 

For the aggregated hair token $\mathbf{T}_{\text{hair}}$, we first decode it into $S = 256$ direction vectors and corresponding vertex positions for each strand, as described in Eq.~\eqref{eq:hair-dir-decode}. From this, we can compute the average strand length ${\lVert \overline{\mathbf{d}} \rVert_2}$, classifying it into short, medium, and long hair categories. 
Then we apply the following steps: (1) The last dimension of the raw DiffLocks~\cite{corr2025difflocks} feature $f_{\text{hair}}$ is treated as a coarse density map, which is scaled and adjusted under different hair lengths. A density-based mask is then used to randomly sample and remove excess strands, effectively reducing the total number of strands in a scalp-aware manner;
(2) Based on predefined hyperparameters for base vertices number $S_0=24$ and base radius $r_0$, we use a piecewise linear relationship to compute new vertex count $S'$ and Gaussian radius $r'$. Sampling vertices at intervals reduces the number of Gaussians per strand. It is noted that the strand Gaussian radius increases linearly with the average strand length for short hair, maintaining scalp coverage, though there is no strict growth relationship between categories of different lengths.

The computation is performed at the full UV map scale. While patch-scale adaption provides more refined regional control, the improvement is minimal and comes at the cost of increased forward time. 

\subsection{Experiment Details}

\paragraph{Data Selection and Augmentation.}

During training, for each sample, we randomly select between 1 and 6 images with different views and expressions as input, and 4 images with the same expression as supervision. This means that within a single iteration, the input images have varying expressions, while the supervision images maintain the same expression, allowing the Gaussians reconstructed from casual captures to require only one pose and expression transformation per iteration. To enhance model generalization, we apply random data augmentations, including small perturbations to brightness, contrast, and saturation. The probability of triggering augmentations is set to 0.6, with upper and lower thresholds of 0.2, 0.15, and 0.15 respectively.

During the optional refinement stage, we perform fine-tuning and supervision only on the few input images. The intermediate results from encoding and the forward pass of the transformer backbone are saved and fixed, with subsequent fine-tuning running the bidirectional forward pass only on the Gaussian decoder and renderer. For cases with more than 6 images, we randomly select multiple sets of 6 images, calculate and save the intermediate results for each set, ensuring that all images are used.

\paragraph{Training Setup.}

Our model predicts a set of planar or strand-based Gaussians for each UV pixel. The total number of Gaussians can be controlled by adjusting the size of the texture map. In our experiments, the head UV has dimensions $H_{\text{uv}}^{\text{head}} = W_{\text{uv}}^{\text{head}} = 224$, while the hair scalp UV has dimensions $H_{\text{uv}}^{\text{scalp}} = W_{\text{uv}}^{\text{scalp}} = 112$.

For training, we employ the Distributed Data Parallel (DDP) strategy for multi-GPU training, with L2 norm gradient clipping set to 1 for all learnable parameters in each iteration. A cosine learning rate scheduler is used during the training phase with 600 warm-up iterations, while a linear learning rate scheduler with a decay factor of 0.1 is applied during the fine-tuning phase for 100 epochs. The initial learning rate for both phases is set to 1e-4.

\section{More Results}
\label{sec:sup-more-results}

\paragraph{Results on Tracking Improvements.}

\cref{fig:preprocess-ablation} shows the improvement in FLAME tracking due to lip alignment loss and motion-aware iteration. The test images are monocular data captured by mobile phones from the internet. The addition of lip alignment loss ensures that the lips of the tracked FLAME model close accurately when the person closes their mouth, while the motion-aware iteration schedule improves the shape accuracy, better aligning with the facial and lip shapes. This provides a good prior for avatar reconstruction and new expression reenactment.

\paragraph{Results on Hair–Face Disentanglement.} 

We separate face and hair using semantic masks~\cite{eccv2024sapiens} and report region-wise results under 6-input setting in~\cref{table:mask-evaluation}. Our method achieves the best PSNR, SSIM, and LPIPS for each region independently. Moreover, we apply face parsing to both the rendered results and the ground truth and compute region-wise IoU, where our method achieves 0.92 (\(\uparrow 6.7\%\)) and 0.83 (\(\uparrow 36\%\)) for the face and hair regions, respectively, validating the effectiveness of our hair-face disentanglement.

\input{tables/supp_mask_eval}

\paragraph{Challenging Results.}

\begin{figure}[h]
   \centering
   \vspace{-0.5em}
   \includegraphics[width=0.9\linewidth]{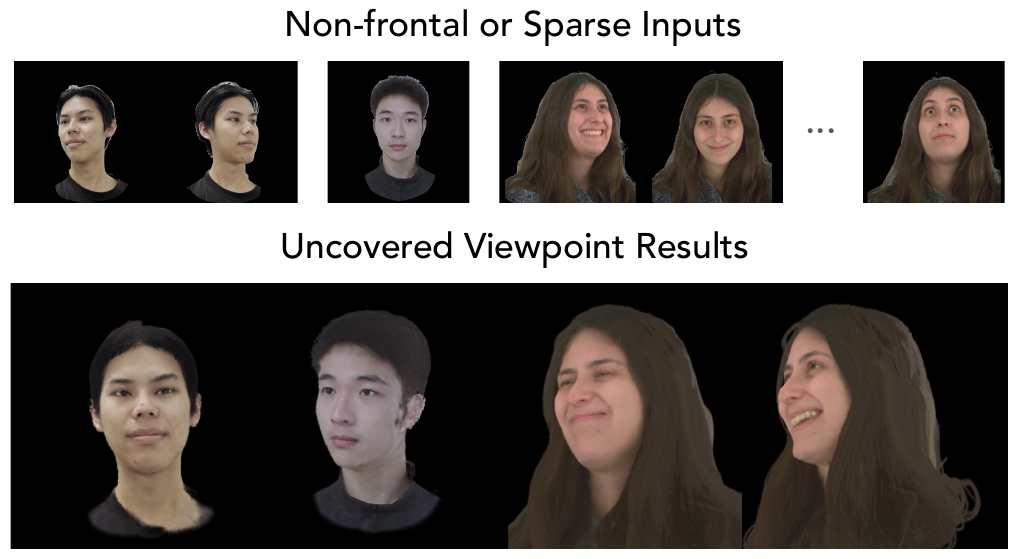}
   \vspace{-0.5em}
   \caption{\textbf{Results under challenging inputs and viewpoints}}
   \vspace{-1.0em}
   \label{fig:more-results-challenging}
\end{figure}

Additional rendering results of uncovered novel viewpoints under more challenging settings, including cases without near-frontal input and with sparse input, are shown in~\cref{fig:more-results-challenging}, further demonstrating the robustness of the proposed method.

\paragraph{Adaptive Hair Branch}

\begin{figure}[h]
    \centering
    \includegraphics[width=\linewidth]{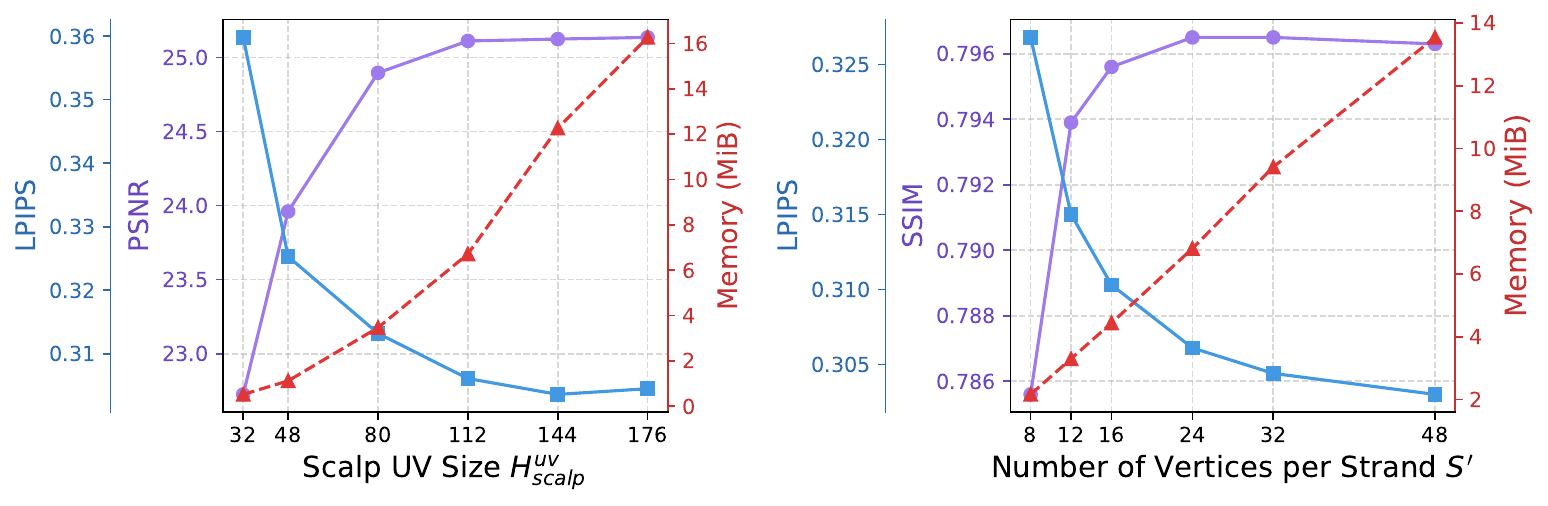}
    \vspace{-1.5em}
    \caption{\textbf{Quality-Efficiency Trade-off} in the Hair Branch.}\label{fig:adaptive-hair-sensitivity}
    \vspace{-0.5em}
\end{figure}

We conduct additional experiments on the adaptive hair branch, starting with parameter selection.The number of hair Gaussians $N$ is jointly determined by the scalp UV resolution $H_{\text{scalp}}^{\mathrm{uv}}$ and the per-strand segment count $S' \propto S_0$, with detailed definitions in Sec.~3.1.3 and App. A.2. We have supplemented parameter sensitivity analysis. As shown in~\cref{fig:adaptive-hair-sensitivity}, we find that $H_{\text{scalp}}^{\mathrm{uv}}=112$ and $S_0=24$ strike a favorable balance between visual quality and memory usage.

\begin{figure}[h]
   \centering
   \vspace{-0.5em}
   \includegraphics[width=0.9\linewidth]{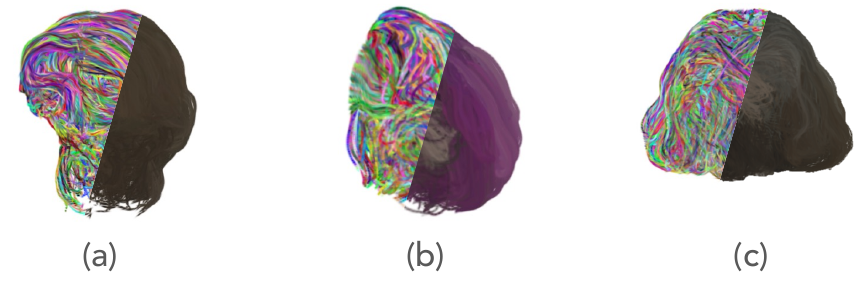}
   \vspace{-0.5em}
   \caption{\textbf{Cases of Different Hairstyles}: (a) male curly hair, (b) female dyed long hair, (c) elderly female medium-length hair.}
   \label{fig:hair-cases}
   \vspace{-1.0em}
\end{figure}

\begin{figure}[h]
    \centering
    \includegraphics[width=0.99\linewidth]{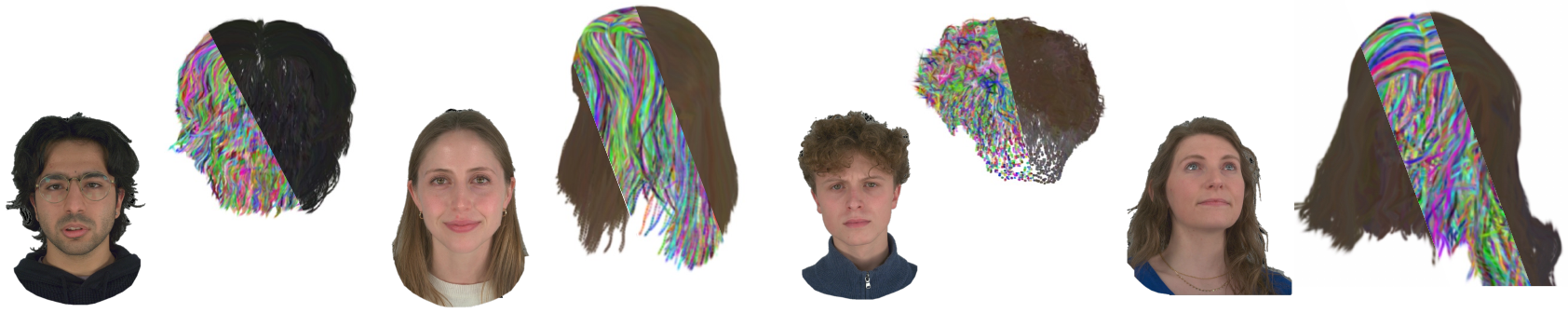}
    \vspace{-0.5em}
    \caption{\textbf{Cases of Challenging Hairstyles.}}
    \label{fig:more-hairstyle}
    \vspace{-1.2em}
\end{figure}

Moreover, \cref{table:adaptive_hair} explicitly reports the hair-sampling data for the three individual cases shown in~\cref{fig:hair-cases}, where longer hair is expected to correspond to a larger number of Gaussians. \cref{fig:more-hairstyle} further provides results on long, wavy, and curly hairstyles, demonstrating the robustness of our hair-branch design.

\input{tables/supp_hair}

\paragraph{Video Presentation.}

We also provide additional results of \paperName in the \textbf{attached video}, including more renderings of new identities, expressions, and viewpoints on both the NeRSemble dataset~\cite{tog2023nersemble} and real-world in-the-wild data. We also visualize the same generated avatar under different background colors to ensure objectivity and introduce greater variability.

\section{Additional Discussion}

\paragraph{Limitations and Future Work.}

Since our reconstruction and animation of 3D Gaussian avatars are based on the FLAME model and its parameters, it remains difficult to represent static regions and dynamic details that are not modeled by FLAME, such as the tongue and fine facial wrinkles. In addition, although we achieve strand-level modeling for hair, our training set still exhibits bias in hairstyle distribution, leading to occasional failures on complex accessories or uncommon hairstyles.

For future work, the generated Gaussian avatars can be exported to TaoAvatar's~\cite{cvpr2025taoavatar} optimized renderer, enabling efficient rendering and Text-to-Speech (TTS) integration on mobile devices such as Apple Vision Pro. Such integration would further support applications including online meetings, virtual companions, and VR gaming, providing a high-quality foundation for interactive digital experiences.

\paragraph{Potential Social Impact.}
\paperName can generate realistic 3D head avatars and synthetic renderings from casual captures. 
As with other generative methods capable of producing lifelike digital humans, responsible use is advised to avoid unintended misuse or misrepresentation.

\end{appendix}

%% file: tables/supp_mask_eval.tex
\begin{table*}[!t]
    \caption{\textbf{Face-Region and Hair-Region Evaluation Results.}}\label{table:mask-evaluation}
    \vspace{-0.5em}
    \centering
    \begin{tabular}{l|c|cccc|cccc}
        \toprule
        & \multicolumn{1}{c|}{Overall} 
        & \multicolumn{4}{c|}{Face (w/o Neck)} 
        & \multicolumn{4}{c}{Hair} \\
        Method 
        & PSNR$\uparrow$ 
        & PSNR$\uparrow$ & SSIM$\uparrow$ & LPIPS$\downarrow$ & IoU$\uparrow$
        & PSNR$\uparrow$ & SSIM$\uparrow$ & LPIPS$\downarrow$ & IoU$\uparrow$ \\
        \midrule
        DiffusionRig~\cite{cvpr2023diffusionrig}   &16.55& 13.78& 0.671& 0.359& 0.506& 17.50& 0.807&  0.216& 0.419\\
        FlashAvatar~\cite{cvpr2024flashavatar}    &16.08& 14.64& 0.651& 0.336& 0.220& 16.91& 0.773&  0.210& 0.179\\
        GaussianAvatars~\cite{cvpr2024gaussianavatars}&\underline{23.44}& \underline{21.48}& \underline{0.784}& \underline{0.200}& \underline{0.864}& \underline{23.54}& \underline{0.830}&  \underline{0.163}& \underline{0.608}\\
        MeGA~\cite{cvpr2025mega}          &17.29& 16.92& 0.649& 0.302& 0.624& 19.60& 0.759&  0.193& 0.476\\
        Ours                &\textbf{23.71}& \textbf{24.60}& \textbf{0.880}& \textbf{0.102}& \textbf{0.922}& \textbf{24.46}& \textbf{0.867}& \textbf{0.157}& \textbf{0.826}\\
        \bottomrule
    \end{tabular}
    \vspace{-0.5em}
\end{table*}

%% file: tables/supp_hair.tex
\begin{table}[h]
    \caption{{\bf Adaptive Hair Gaussians}.}
    \vspace{-0.5em}
    \centering
    \scalebox{1.0}{
        \begin{tabular}{c|ccc|c}
        \toprule
        Case & ${\lVert \overline{\mathbf{d}} \rVert_2}$ & Density Scale & $S'$ & Total Gaussians\\ \midrule
        (a) & 0.121 & 1.235 & 21 & 54080\\
        (b) & 0.228 & 1.000 & 42 & 84132\\
        (c) & 0.201 & 1.000 & 36 & 77455\\
        \bottomrule
        \end{tabular}}
    \label{table:adaptive_hair}
\end{table}

%% file: main.bib
@String(CVPR= {IEEE Conf. Comput. Vis. Pattern Recog.})

@String(ICCV= {Int. Conf. Comput. Vis.})

@String(ECCV= {Eur. Conf. Comput. Vis.})

@String(TOG= {ACM Trans. Graph.})

@String(ICASSP=	{ICASSP})

@String(ICLR = {Int. Conf. Learn. Represent.})

@String(CVPR  = {CVPR})

@String(ICCV  = {ICCV})

@String(ECCV  = {ECCV})

@String(TOG   = {ACM TOG})

@String(ICLR  = {ICLR})

@article{tog20233dgaussian,
  author       = {Bernhard Kerbl and
                  Georgios Kopanas and
                  Thomas Leimk{\"{u}}hler and
                  George Drettakis},
  title        = {3D Gaussian Splatting for Real-Time Radiance Field Rendering},
  journal      = {{ACM} Trans. Graph.},
  volume       = {42},
  number       = {4},
  pages        = {139:1--139:14},
  year         = {2023},
  url          = {https://doi.org/10.1145/3592433},
  doi          = {10.1145/3592433},
  bibsource    = {dblp computer science bibliography, https://dblp.org}
}

@article{tog2017flame,
  author       = {Tianye Li and
                  Timo Bolkart and
                  Michael J. Black and
                  Hao Li and
                  Javier Romero},
  title        = {Learning a model of facial shape and expression from 4D scans},
  journal      = {{ACM} Trans. Graph.},
  volume       = {36},
  number       = {6},
  pages        = {194:1--194:17},
  year         = {2017},
  url          = {https://doi.org/10.1145/3130800.3130813},
  doi          = {10.1145/3130800.3130813},
  bibsource    = {dblp computer science bibliography, https://dblp.org}
}

@inproceedings{corr2025vggt,
  author       = {Jianyuan Wang and
                  Minghao Chen and
                  Nikita Karaev and
                  Andrea Vedaldi and
                  Christian Rupprecht and
                  David Novotn{\'{y}}},
  title        = {{VGGT:} Visual Geometry Grounded Transformer},
  booktitle    = {{IEEE/CVF} Conference on Computer Vision and Pattern Recognition,
                  {CVPR} 2025, Nashville, TN, USA, June 11-15, 2025},
  pages        = {5294--5306},
  publisher    = {Computer Vision Foundation / {IEEE}},
  year         = {2025},
  url          = {https://openaccess.thecvf.com/content/CVPR2025/html/Wang\_VGGT\_Visual\_Geometry\_Grounded\_Transformer\_CVPR\_2025\_paper.html},
  doi          = {10.1109/CVPR52734.2025.00499},
  bibsource    = {dblp computer science bibliography, https://dblp.org}
}

@inproceedings{cvpr2018perceploss,
  author       = {Richard Zhang and
                  Phillip Isola and
                  Alexei A. Efros and
                  Eli Shechtman and
                  Oliver Wang},
  title        = {The Unreasonable Effectiveness of Deep Features as a Perceptual Metric},
  booktitle    = {2018 {IEEE} Conference on Computer Vision and Pattern Recognition,
                  {CVPR} 2018, Salt Lake City, UT, USA, June 18-22, 2018},
  pages        = {586--595},
  publisher    = {Computer Vision Foundation / {IEEE} Computer Society},
  year         = {2018},
  url          = {http://openaccess.thecvf.com/content\_cvpr\_2018/html/Zhang\_The\_Unreasonable\_Effectiveness\_CVPR\_2018\_paper.html},
  doi          = {10.1109/CVPR.2018.00068},
  bibsource    = {dblp computer science bibliography, https://dblp.org}
}

@inproceedings{cvpr2024gaussianavatars,
  author       = {Shenhan Qian and
                  Tobias Kirschstein and
                  Liam Schoneveld and
                  Davide Davoli and
                  Simon Giebenhain and
                  Matthias Nie{\ss}ner},
  title        = {GaussianAvatars: Photorealistic Head Avatars with Rigged 3D Gaussians},
  booktitle    = {{IEEE/CVF} Conference on Computer Vision and Pattern Recognition,
                  {CVPR} 2024, Seattle, WA, USA, June 16-22, 2024},
  pages        = {20299--20309},
  publisher    = {{IEEE}},
  year         = {2024},
  url          = {https://doi.org/10.1109/CVPR52733.2024.01919},
  doi          = {10.1109/CVPR52733.2024.01919},
  bibsource    = {dblp computer science bibliography, https://dblp.org}
}

@inproceedings{cvpr2021bgmatting,
  author       = {Shanchuan Lin and
                  Andrey Ryabtsev and
                  Soumyadip Sengupta and
                  Brian L. Curless and
                  Steven M. Seitz and
                  Ira Kemelmacher{-}Shlizerman},
  title        = {Real-Time High-Resolution Background Matting},
  booktitle    = {{IEEE} Conference on Computer Vision and Pattern Recognition, {CVPR}
                  2021, virtual, June 19-25, 2021},
  pages        = {8762--8771},
  publisher    = {Computer Vision Foundation / {IEEE}},
  year         = {2021},
  url          = {https://openaccess.thecvf.com/content/CVPR2021/html/Lin\_Real-Time\_High-Resolution\_Background\_Matting\_CVPR\_2021\_paper.html},
  doi          = {10.1109/CVPR46437.2021.00865},
  bibsource    = {dblp computer science bibliography, https://dblp.org}
}

@article{tog2023nersemble,
  author       = {Tobias Kirschstein and
                  Shenhan Qian and
                  Simon Giebenhain and
                  Tim Walter and
                  Matthias Nie{\ss}ner},
  title        = {NeRSemble: Multi-view Radiance Field Reconstruction of Human Heads},
  journal      = TOG,
  volume       = {42},
  number       = {4},
  pages        = {161:1--161:14},
  year         = {2023},
  url          = {https://doi.org/10.1145/3592455},
  doi          = {10.1145/3592455},
  bibsource    = {dblp computer science bibliography, https://dblp.org}
}

@inproceedings{cvpr2022vfhq,
  author       = {Liangbin Xie and
                  Xintao Wang and
                  Honglun Zhang and
                  Chao Dong and
                  Ying Shan},
  title        = {{VFHQ:} {A} High-Quality Dataset and Benchmark for Video Face Super-Resolution},
  booktitle    = {{IEEE/CVF} Conference on Computer Vision and Pattern Recognition Workshops,
                  {CVPR} Workshops 2022, New Orleans, LA, USA, June 19-20, 2022},
  pages        = {656--665},
  publisher    = {{IEEE}},
  year         = {2022},
  url          = {https://doi.org/10.1109/CVPRW56347.2022.00081},
  doi          = {10.1109/CVPRW56347.2022.00081},
  bibsource    = {dblp computer science bibliography, https://dblp.org}
}

@article{tip2004ssim,
  author       = {Zhou Wang and
                  Alan C. Bovik and
                  Hamid R. Sheikh and
                  Eero P. Simoncelli},
  title        = {Image quality assessment: from error visibility to structural similarity},
  journal      = {{IEEE} Trans. Image Process.},
  volume       = {13},
  number       = {4},
  pages        = {600--612},
  year         = {2004},
  url          = {https://doi.org/10.1109/TIP.2003.819861},
  doi          = {10.1109/TIP.2003.819861},
  bibsource    = {dblp computer science bibliography, https://dblp.org}
}

@inproceedings{cvpr2018lpips,
  author       = {Richard Zhang and
                  Phillip Isola and
                  Alexei A. Efros and
                  Eli Shechtman and
                  Oliver Wang},
  title        = {The Unreasonable Effectiveness of Deep Features as a Perceptual Metric},
  booktitle    = {2018 {IEEE} Conference on Computer Vision and Pattern Recognition,
                  {CVPR} 2018, Salt Lake City, UT, USA, June 18-22, 2018},
  pages        = {586--595},
  publisher    = {Computer Vision Foundation / {IEEE} Computer Society},
  year         = {2018},
  url          = {http://openaccess.thecvf.com/content\_cvpr\_2018/html/Zhang\_The\_Unreasonable\_Effectiveness\_CVPR\_2018\_paper.html},
  doi          = {10.1109/CVPR.2018.00068},
  bibsource    = {dblp computer science bibliography, https://dblp.org}
}

@inproceedings{eccv2020nerf,
  author       = {Ben Mildenhall and
                  Pratul P. Srinivasan and
                  Matthew Tancik and
                  Jonathan T. Barron and
                  Ravi Ramamoorthi and
                  Ren Ng},
  editor       = {Andrea Vedaldi and
                  Horst Bischof and
                  Thomas Brox and
                  Jan{-}Michael Frahm},
  title        = {NeRF: Representing Scenes as Neural Radiance Fields for View Synthesis},
  booktitle    = {Computer Vision - {ECCV} 2020 - 16th European Conference, Glasgow,
                  UK, August 23-28, 2020, Proceedings, Part {I}},
  series       = {Lecture Notes in Computer Science},
  volume       = {12346},
  pages        = {405--421},
  publisher    = {Springer},
  year         = {2020},
  url          = {https://doi.org/10.1007/978-3-030-58452-8\_24},
  doi          = {10.1007/978-3-030-58452-8\_24},
  bibsource    = {dblp computer science bibliography, https://dblp.org}
}

@article{tmlr2024dinov2,
  author       = {Maxime Oquab and
                  Timoth{\'{e}}e Darcet and
                  Th{\'{e}}o Moutakanni and
                  Huy V. Vo and
                  Marc Szafraniec and
                  Vasil Khalidov and
                  Pierre Fernandez and
                  Daniel Haziza and
                  Francisco Massa and
                  Alaaeldin El{-}Nouby and
                  Mido Assran and
                  Nicolas Ballas and
                  Wojciech Galuba and
                  Russell Howes and
                  Po{-}Yao Huang and
                  Shang{-}Wen Li and
                  Ishan Misra and
                  Michael Rabbat and
                  Vasu Sharma and
                  Gabriel Synnaeve and
                  Hu Xu and
                  Herv{\'{e}} J{\'{e}}gou and
                  Julien Mairal and
                  Patrick Labatut and
                  Armand Joulin and
                  Piotr Bojanowski},
  title        = {DINOv2: Learning Robust Visual Features without Supervision},
  journal      = {Trans. Mach. Learn. Res.},
  volume       = {2024},
  year         = {2024},
  url          = {https://openreview.net/forum?id=a68SUt6zFt},
  bibsource    = {dblp computer science bibliography, https://dblp.org}
}

@inproceedings{iccv2021dpt,
  author       = {Ren{\'{e}} Ranftl and
                  Alexey Bochkovskiy and
                  Vladlen Koltun},
  title        = {Vision Transformers for Dense Prediction},
  booktitle    = {2021 {IEEE/CVF} International Conference on Computer Vision, {ICCV}
                  2021, Montreal, QC, Canada, October 10-17, 2021},
  pages        = {12159--12168},
  publisher    = {{IEEE}},
  year         = {2021},
  url          = {https://doi.org/10.1109/ICCV48922.2021.01196},
  doi          = {10.1109/ICCV48922.2021.01196},
  bibsource    = {dblp computer science bibliography, https://dblp.org}
}

@inproceedings{iclr2015adam,
  author       = {Diederik P. Kingma and
                  Jimmy Ba},
  editor       = {Yoshua Bengio and
                  Yann LeCun},
  title        = {Adam: {A} Method for Stochastic Optimization},
  booktitle    = {3rd International Conference on Learning Representations, {ICLR} 2015,
                  San Diego, CA, USA, May 7-9, 2015, Conference Track Proceedings},
  year         = {2015},
  url          = {http://arxiv.org/abs/1412.6980},
  bibsource    = {dblp computer science bibliography, https://dblp.org}
}

@inproceedings{mm2024gaussiantalker,
  author       = {Hongyun Yu and
                  Zhan Qu and
                  Qihang Yu and
                  Jianchuan Chen and
                  Zhonghua Jiang and
                  Zhiwen Chen and
                  Shengyu Zhang and
                  Jimin Xu and
                  Fei Wu and
                  Chengfei Lv and
                  Gang Yu},
  editor       = {Jianfei Cai and
                  Mohan S. Kankanhalli and
                  Balakrishnan Prabhakaran and
                  Susanne Boll and
                  Ramanathan Subramanian and
                  Liang Zheng and
                  Vivek K. Singh and
                  Pablo C{\'{e}}sar and
                  Lexing Xie and
                  Dong Xu},
  title        = {GaussianTalker: Speaker-specific Talking Head Synthesis via 3D Gaussian
                  Splatting},
  booktitle    = {Proceedings of the 32nd {ACM} International Conference on Multimedia,
                  {MM} 2024, Melbourne, VIC, Australia, 28 October 2024 - 1 November
                  2024},
  pages        = {3548--3557},
  publisher    = {{ACM}},
  year         = {2024},
  url          = {https://doi.org/10.1145/3664647.3681675},
  doi          = {10.1145/3664647.3681675},
  bibsource    = {dblp computer science bibliography, https://dblp.org}
}

@article{corr2025difflocks,
  author       = {Radu Alexandru Rosu and
                  Keyu Wu and
                  Yao Feng and
                  Youyi Zheng and
                  Michael J. Black},
  title        = {DiffLocks: Generating 3D Hair from a Single Image using Diffusion
                  Models},
  journal      = {CoRR},
  volume       = {abs/2505.06166},
  year         = {2025},
  url          = {https://doi.org/10.48550/arXiv.2505.06166},
  doi          = {10.48550/ARXIV.2505.06166},
  eprinttype    = {arXiv},
  eprint       = {2505.06166},
  bibsource    = {dblp computer science bibliography, https://dblp.org}
}

@inproceedings{iccv2021siren,
  author       = {Ishit Mehta and
                  Micha{\"{e}}l Gharbi and
                  Connelly Barnes and
                  Eli Shechtman and
                  Ravi Ramamoorthi and
                  Manmohan Chandraker},
  title        = {Modulated Periodic Activations for Generalizable Local Functional
                  Representations},
  booktitle    = {2021 {IEEE/CVF} International Conference on Computer Vision, {ICCV}
                  2021, Montreal, QC, Canada, October 10-17, 2021},
  pages        = {14194--14203},
  publisher    = {{IEEE}},
  year         = {2021},
  url          = {https://doi.org/10.1109/ICCV48922.2021.01395},
  doi          = {10.1109/ICCV48922.2021.01395},
  bibsource    = {dblp computer science bibliography, https://dblp.org}
}

@inproceedings{eccv2024sapiens,
  author       = {Rawal Khirodkar and
                  Timur M. Bagautdinov and
                  Julieta Martinez and
                  Su Zhaoen and
                  Austin James and
                  Peter Selednik and
                  Stuart Anderson and
                  Shunsuke Saito},
  editor       = {Ales Leonardis and
                  Elisa Ricci and
                  Stefan Roth and
                  Olga Russakovsky and
                  Torsten Sattler and
                  G{\"{u}}l Varol},
  title        = {Sapiens: Foundation for Human Vision Models},
  booktitle    = {Computer Vision - {ECCV} 2024 - 18th European Conference, Milan, Italy,
                  September 29-October 4, 2024, Proceedings, Part {IV}},
  series       = {Lecture Notes in Computer Science},
  volume       = {15062},
  pages        = {206--228},
  publisher    = {Springer},
  year         = {2024},
  url          = {https://doi.org/10.1007/978-3-031-73235-5\_12},
  doi          = {10.1007/978-3-031-73235-5\_12},
  bibsource    = {dblp computer science bibliography, https://dblp.org}
}

@inproceedings{siggraph19993dmm,
  author       = {Volker Blanz and
                  Thomas Vetter},
  editor       = {Warren N. Waggenspack},
  title        = {A Morphable Model for the Synthesis of 3D Faces},
  booktitle    = {Proceedings of the 26th Annual Conference on Computer Graphics and
                  Interactive Techniques, {SIGGRAPH} 1999, Los Angeles, CA, USA, August
                  8-13, 1999},
  pages        = {187--194},
  publisher    = {{ACM}},
  year         = {1999},
  url          = {https://dl.acm.org/citation.cfm?id=311556},
  bibsource    = {dblp computer science bibliography, https://dblp.org}
}

@inproceedings{cvpr2019voca,
  author       = {Daniel Cudeiro and
                  Timo Bolkart and
                  Cassidy Laidlaw and
                  Anurag Ranjan and
                  Michael J. Black},
  title        = {Capture, Learning, and Synthesis of 3D Speaking Styles},
  booktitle    = {{IEEE} Conference on Computer Vision and Pattern Recognition, {CVPR}
                  2019, Long Beach, CA, USA, June 16-20, 2019},
  pages        = {10101--10111},
  publisher    = {Computer Vision Foundation / {IEEE}},
  year         = {2019},
  url          = {http://openaccess.thecvf.com/content\_CVPR\_2019/html/Cudeiro\_Capture\_Learning\_and\_Synthesis\_of\_3D\_Speaking\_Styles\_CVPR\_2019\_paper.html},
  doi          = {10.1109/CVPR.2019.01034},
  bibsource    = {dblp computer science bibliography, https://dblp.org}
}

@article{tog2021deca,
  author       = {Yao Feng and
                  Haiwen Feng and
                  Michael J. Black and
                  Timo Bolkart},
  title        = {Learning an animatable detailed 3D face model from in-the-wild images},
  journal      = {{ACM} Trans. Graph.},
  volume       = {40},
  number       = {4},
  pages        = {88:1--88:13},
  year         = {2021},
  url          = {https://doi.org/10.1145/3450626.3459936},
  doi          = {10.1145/3450626.3459936},
  bibsource    = {dblp computer science bibliography, https://dblp.org}
}

@inproceedings{cvpr2022emoca,
  author       = {Radek Danecek and
                  Michael J. Black and
                  Timo Bolkart},
  title        = {{EMOCA:} Emotion Driven Monocular Face Capture and Animation},
  booktitle    = {{IEEE/CVF} Conference on Computer Vision and Pattern Recognition,
                  {CVPR} 2022, New Orleans, LA, USA, June 18-24, 2022},
  pages        = {20279--20290},
  publisher    = {{IEEE}},
  year         = {2022},
  url          = {https://doi.org/10.1109/CVPR52688.2022.01967},
  doi          = {10.1109/CVPR52688.2022.01967},
  bibsource    = {dblp computer science bibliography, https://dblp.org}
}

@article{tog2018deep-video-portaits,
  author       = {Hyeongwoo Kim and
                  Pablo Garrido and
                  Ayush Tewari and
                  Weipeng Xu and
                  Justus Thies and
                  Matthias Nie{\ss}ner and
                  Patrick P{\'{e}}rez and
                  Christian Richardt and
                  Michael Zollh{\"{o}}fer and
                  Christian Theobalt},
  title        = {Deep video portraits},
  journal      = {{ACM} Trans. Graph.},
  volume       = {37},
  number       = {4},
  pages        = {163},
  year         = {2018},
  url          = {https://doi.org/10.1145/3197517.3201283},
  doi          = {10.1145/3197517.3201283},
  bibsource    = {dblp computer science bibliography, https://dblp.org}
}

@inproceedings{cvpr2022nha-from-mono,
  author       = {Philip{-}William Grassal and
                  Malte Prinzler and
                  Titus Leistner and
                  Carsten Rother and
                  Matthias Nie{\ss}ner and
                  Justus Thies},
  title        = {Neural Head Avatars from Monocular {RGB} Videos},
  booktitle    = {{IEEE/CVF} Conference on Computer Vision and Pattern Recognition,
                  {CVPR} 2022, New Orleans, LA, USA, June 18-24, 2022},
  pages        = {18632--18643},
  publisher    = {{IEEE}},
  year         = {2022},
  url          = {https://doi.org/10.1109/CVPR52688.2022.01810},
  doi          = {10.1109/CVPR52688.2022.01810},
  bibsource    = {dblp computer science bibliography, https://dblp.org}
}

@inproceedings{cvpr2022rignerf,
  author       = {ShahRukh Athar and
                  Zexiang Xu and
                  Kalyan Sunkavalli and
                  Eli Shechtman and
                  Zhixin Shu},
  title        = {RigNeRF: Fully Controllable Neural 3D Portraits},
  booktitle    = {{IEEE/CVF} Conference on Computer Vision and Pattern Recognition,
                  {CVPR} 2022, New Orleans, LA, USA, June 18-24, 2022},
  pages        = {20332--20341},
  publisher    = {{IEEE}},
  year         = {2022},
  url          = {https://doi.org/10.1109/CVPR52688.2022.01972},
  doi          = {10.1109/CVPR52688.2022.01972},
  bibsource    = {dblp computer science bibliography, https://dblp.org}
}

@inproceedings{cvpr2023avatar-from-monocular,
  author       = {Ziqian Bai and
                  Feitong Tan and
                  Zeng Huang and
                  Kripasindhu Sarkar and
                  Danhang Tang and
                  Di Qiu and
                  Abhimitra Meka and
                  Ruofei Du and
                  Mingsong Dou and
                  Sergio Orts{-}Escolano and
                  Rohit Pandey and
                  Ping Tan and
                  Thabo Beeler and
                  Sean Fanello and
                  Yinda Zhang},
  title        = {Learning Personalized High Quality Volumetric Head Avatars from Monocular
                  {RGB} Videos},
  booktitle    = {{IEEE/CVF} Conference on Computer Vision and Pattern Recognition,
                  {CVPR} 2023, Vancouver, BC, Canada, June 17-24, 2023},
  pages        = {16890--16900},
  publisher    = {{IEEE}},
  year         = {2023},
  url          = {https://doi.org/10.1109/CVPR52729.2023.01620},
  doi          = {10.1109/CVPR52729.2023.01620},
  bibsource    = {dblp computer science bibliography, https://dblp.org}
}

@inproceedings{cvpr2021nerface,
  author       = {Guy Gafni and
                  Justus Thies and
                  Michael Zollh{\"{o}}fer and
                  Matthias Nie{\ss}ner},
  title        = {Dynamic Neural Radiance Fields for Monocular 4D Facial Avatar Reconstruction},
  booktitle    = {{IEEE} Conference on Computer Vision and Pattern Recognition, {CVPR}
                  2021, virtual, June 19-25, 2021},
  pages        = {8649--8658},
  publisher    = {Computer Vision Foundation / {IEEE}},
  year         = {2021},
  url          = {https://openaccess.thecvf.com/content/CVPR2021/html/Gafni\_Dynamic\_Neural\_Radiance\_Fields\_for\_Monocular\_4D\_Facial\_Avatar\_Reconstruction\_CVPR\_2021\_paper.html},
  doi          = {10.1109/CVPR46437.2021.00854},
  bibsource    = {dblp computer science bibliography, https://dblp.org}
}

@inproceedings{cvpr2021headnerf,
     author     = {Yang Hong and Bo Peng and Haiyao Xiao and Ligang Liu and Juyong Zhang},
     title      = {HeadNeRF: A Real-time NeRF-based Parametric Head Model},
     booktitle  = {{IEEE/CVF} Conference on Computer Vision and Pattern Recognition (CVPR)},
     year       = {2022}
  }

@inproceedings{fg2023flame-in-nerf,
  author       = {ShahRukh Athar and
                  Zhixin Shu and
                  Dimitris Samaras},
  title        = {FLAME-in-NeRF: Neural control of Radiance Fields for Free View Face
                  Animation},
  booktitle    = {17th {IEEE} International Conference on Automatic Face and Gesture
                  Recognition, {FG} 2023, Waikoloa Beach, HI, USA, January 5-8, 2023},
  pages        = {1--8},
  publisher    = {{IEEE}},
  year         = {2023},
  url          = {https://doi.org/10.1109/FG57933.2023.10042553},
  doi          = {10.1109/FG57933.2023.10042553},
  bibsource    = {dblp computer science bibliography, https://dblp.org}
}

@inproceedings{cvpr2023insta,
  author       = {Wojciech Zielonka and
                  Timo Bolkart and
                  Justus Thies},
  title        = {Instant Volumetric Head Avatars},
  booktitle    = {{IEEE/CVF} Conference on Computer Vision and Pattern Recognition,
                  {CVPR} 2023, Vancouver, BC, Canada, June 17-24, 2023},
  pages        = {4574--4584},
  publisher    = {{IEEE}},
  year         = {2023},
  url          = {https://doi.org/10.1109/CVPR52729.2023.00444},
  doi          = {10.1109/CVPR52729.2023.00444},
  bibsource    = {dblp computer science bibliography, https://dblp.org}
}

@inproceedings{cvpr2024flashavatar,
  author       = {Jun Xiang and
                  Xuan Gao and
                  Yudong Guo and
                  Juyong Zhang},
  title        = {FlashAvatar: High-Fidelity Head Avatar with Efficient Gaussian Embedding},
  booktitle    = {{IEEE/CVF} Conference on Computer Vision and Pattern Recognition,
                  {CVPR} 2024, Seattle, WA, USA, June 16-22, 2024},
  pages        = {1802--1812},
  publisher    = {{IEEE}},
  year         = {2024},
  url          = {https://doi.org/10.1109/CVPR52733.2024.00177},
  doi          = {10.1109/CVPR52733.2024.00177},
  bibsource    = {dblp computer science bibliography, https://dblp.org}
}

@inproceedings{eccv2024HeadGaS,
  author       = {Helisa Dhamo and
                  Yinyu Nie and
                  Arthur Moreau and
                  Jifei Song and
                  Richard Shaw and
                  Yiren Zhou and
                  Eduardo P{\'{e}}rez{-}Pellitero},
  editor       = {Ales Leonardis and
                  Elisa Ricci and
                  Stefan Roth and
                  Olga Russakovsky and
                  Torsten Sattler and
                  G{\"{u}}l Varol},
  title        = {HeadGaS: Real-Time Animatable Head Avatars via 3D Gaussian Splatting},
  booktitle    = {Computer Vision - {ECCV} 2024 - 18th European Conference, Milan, Italy,
                  September 29-October 4, 2024, Proceedings, Part {II}},
  series       = {Lecture Notes in Computer Science},
  volume       = {15060},
  pages        = {459--476},
  publisher    = {Springer},
  year         = {2024},
  url          = {https://doi.org/10.1007/978-3-031-72627-9\_26},
  doi          = {10.1007/978-3-031-72627-9\_26},
  bibsource    = {dblp computer science bibliography, https://dblp.org}
}

@inproceedings{siggragh2024monogaussianavatar,
  author       = {Yufan Chen and
                  Lizhen Wang and
                  Qijing Li and
                  Hongjiang Xiao and
                  Shengping Zhang and
                  Hongxun Yao and
                  Yebin Liu},
  editor       = {Andres Burbano and
                  Denis Zorin and
                  Wojciech Jarosz},
  title        = {MonoGaussianAvatar: Monocular Gaussian Point-based Head Avatar},
  booktitle    = {{ACM} {SIGGRAPH} 2024 Conference Papers, {SIGGRAPH} 2024, Denver,
                  CO, USA, 27 July 2024- 1 August 2024},
  pages        = {58},
  publisher    = {{ACM}},
  year         = {2024},
  url          = {https://doi.org/10.1145/3641519.3657499},
  doi          = {10.1145/3641519.3657499},
  bibsource    = {dblp computer science bibliography, https://dblp.org}
}

@article{tvcg2025gaussianhead,
  author       = {Jie Wang and
                  Jiucheng Xie and
                  Xianyan Li and
                  Feng Xu and
                  Chi{-}Man Pun and
                  Hao Gao},
  title        = {GaussianHead: High-Fidelity Head Avatars With Learnable Gaussian Derivation},
  journal      = {{IEEE} Trans. Vis. Comput. Graph.},
  volume       = {31},
  number       = {7},
  pages        = {4141--4154},
  year         = {2025},
  url          = {https://doi.org/10.1109/TVCG.2025.3561794},
  doi          = {10.1109/TVCG.2025.3561794},
  bibsource    = {dblp computer science bibliography, https://dblp.org}
}

@article{tog2022ava,
  author       = {Chen Cao and
                  Tomas Simon and
                  Jin Kyu Kim and
                  Gabe Schwartz and
                  Michael Zollh{\"{o}}fer and
                  Shunsuke Saito and
                  Stephen Lombardi and
                  Shih{-}En Wei and
                  Danielle Belko and
                  Shoou{-}I Yu and
                  Yaser Sheikh and
                  Jason M. Saragih},
  title        = {Authentic volumetric avatars from a phone scan},
  journal      = {{ACM} Trans. Graph.},
  volume       = {41},
  number       = {4},
  pages        = {163:1--163:19},
  year         = {2022},
  url          = {https://doi.org/10.1145/3528223.3530143},
  doi          = {10.1145/3528223.3530143},
  bibsource    = {dblp computer science bibliography, https://dblp.org}
}

@inproceedings{iccv2023preface,
  author       = {Marcel C. B{\"{u}}hler and
                  Kripasindhu Sarkar and
                  Tanmay Shah and
                  Gengyan Li and
                  Daoye Wang and
                  Leonhard Helminger and
                  Sergio Orts{-}Escolano and
                  Dmitry Lagun and
                  Otmar Hilliges and
                  Thabo Beeler and
                  Abhimitra Meka},
  title        = {Preface: {A} Data-driven Volumetric Prior for Few-shot Ultra High-resolution
                  Face Synthesis},
  booktitle    = {{IEEE/CVF} International Conference on Computer Vision, {ICCV} 2023,
                  Paris, France, October 1-6, 2023},
  pages        = {3379--3390},
  publisher    = {{IEEE}},
  year         = {2023},
  url          = {https://doi.org/10.1109/ICCV51070.2023.00315},
  doi          = {10.1109/ICCV51070.2023.00315},
  bibsource    = {dblp computer science bibliography, https://dblp.org}
}

@inproceedings{3dv2025headgap,
  author       = {Xiaozheng Zheng and
                  Chao Wen and
                  Zhaohu Li and
                  Weiyi Zhang and
                  Zhuo Su and
                  Xu Chang and
                  Yang Zhao and
                  Zheng Lv and
                  Xiaoyuan Zhang and
                  Yongjie Zhang and
                  Guidong Wang and
                  Lan Xu},
  title        = {HeadGAP: Few-Shot 3D Head Avatar via Generalizable Gaussian Priors},
  booktitle    = {International Conference on 3D Vision, 3DV 2025, Singapore, March
                  25-28, 2025},
  pages        = {946--957},
  publisher    = {{IEEE}},
  year         = {2025},
  url          = {https://doi.org/10.1109/3DV66043.2025.00092},
  doi          = {10.1109/3DV66043.2025.00092},
  bibsource    = {dblp computer science bibliography, https://dblp.org}
}

@inproceedings{eccv2024gphm,
  author       = {Yuelang Xu and
                  Lizhen Wang and
                  Zerong Zheng and
                  Zhaoqi Su and
                  Yebin Liu},
  editor       = {Ales Leonardis and
                  Elisa Ricci and
                  Stefan Roth and
                  Olga Russakovsky and
                  Torsten Sattler and
                  G{\"{u}}l Varol},
  title        = {3D Gaussian Parametric Head Model},
  booktitle    = {Computer Vision - {ECCV} 2024 - 18th European Conference, Milan, Italy,
                  September 29-October 4, 2024, Proceedings, Part {XXXV}},
  series       = {Lecture Notes in Computer Science},
  volume       = {15093},
  pages        = {129--147},
  publisher    = {Springer},
  year         = {2024},
  url          = {https://doi.org/10.1007/978-3-031-72761-0\_8},
  doi          = {10.1007/978-3-031-72761-0\_8},
  bibsource    = {dblp computer science bibliography, https://dblp.org}
}

@inproceedings{cvpr2025gasp,
  author       = {Jack R. Saunders and
                  Charlie Hewitt and
                  Yanan Jian and
                  Marek Kowalski and
                  Tadas Baltrusaitis and
                  Yiye Chen and
                  Darren Cosker and
                  Virginia Estellers and
                  Nicholas Gyde and
                  Vinay P. Namboodiri and
                  Benjamin E. Lundell},
  title        = {{GASP:} Gaussian Avatars with Synthetic Priors},
  booktitle    = {{IEEE/CVF} Conference on Computer Vision and Pattern Recognition,
                  {CVPR} 2025, Nashville, TN, USA, June 11-15, 2025},
  pages        = {271--280},
  publisher    = {Computer Vision Foundation / {IEEE}},
  year         = {2025},
  url          = {https://openaccess.thecvf.com/content/CVPR2025/html/Saunders\_GASP\_Gaussian\_Avatars\_with\_Synthetic\_Priors\_CVPR\_2025\_paper.html},
  doi          = {10.1109/CVPR52734.2025.00034},
  bibsource    = {dblp computer science bibliography, https://dblp.org}
}

@inproceedings{nips2024gagavatar,
  author       = {Xuangeng Chu and
                  Tatsuya Harada},
  editor       = {Amir Globersons and
                  Lester Mackey and
                  Danielle Belgrave and
                  Angela Fan and
                  Ulrich Paquet and
                  Jakub M. Tomczak and
                  Cheng Zhang},
  title        = {Generalizable and Animatable Gaussian Head Avatar},
  booktitle    = {Advances in Neural Information Processing Systems 38: Annual Conference
                  on Neural Information Processing Systems 2024, NeurIPS 2024, Vancouver,
                  BC, Canada, December 10 - 15, 2024},
  year         = {2024},
  url          = {http://papers.nips.cc/paper\_files/paper/2024/hash/6a14c7f9fb3f42645cfa6bd5aa446819-Abstract-Conference.html},
  bibsource    = {dblp computer science bibliography, https://dblp.org}
}

@inproceedings{cvpr2025cap4d,
  author       = {Felix Taubner and
                  Ruihang Zhang and
                  Mathieu Tuli and
                  David B. Lindell},
  title        = {{CAP4D:} Creating Animatable 4D Portrait Avatars with Morphable Multi-View
                  Diffusion Models},
  booktitle    = {{IEEE/CVF} Conference on Computer Vision and Pattern Recognition,
                  {CVPR} 2025, Nashville, TN, USA, June 11-15, 2025},
  pages        = {5318--5330},
  publisher    = {Computer Vision Foundation / {IEEE}},
  year         = {2025},
  url          = {https://openaccess.thecvf.com/content/CVPR2025/html/Taubner\_CAP4D\_Creating\_Animatable\_4D\_Portrait\_Avatars\_with\_Morphable\_Multi-View\_Diffusion\_CVPR\_2025\_paper.html},
  doi          = {10.1109/CVPR52734.2025.00501},
  bibsource    = {dblp computer science bibliography, https://dblp.org}
}

@inproceedings{cvpr2025gaf,
  author       = {Jiapeng Tang and
                  Davide Davoli and
                  Tobias Kirschstein and
                  Liam Schoneveld and
                  Matthias Nie{\ss}ner},
  title        = {{GAF:} Gaussian Avatar Reconstruction from Monocular Videos via Multi-view
                  Diffusion},
  booktitle    = {{IEEE/CVF} Conference on Computer Vision and Pattern Recognition,
                  {CVPR} 2025, Nashville, TN, USA, June 11-15, 2025},
  pages        = {5546--5558},
  publisher    = {Computer Vision Foundation / {IEEE}},
  year         = {2025},
  url          = {https://openaccess.thecvf.com/content/CVPR2025/html/Tang\_GAF\_Gaussian\_Avatar\_Reconstruction\_from\_Monocular\_Videos\_via\_Multi-view\_Diffusion\_CVPR\_2025\_paper.html},
  doi          = {10.1109/CVPR52734.2025.00521},
  bibsource    = {dblp computer science bibliography, https://dblp.org}
}

@misc{corr2024facelift,
      title={FaceLift: Single Image to 3D Head with View Generation and GS-LRM}, 
      author={Weijie Lyu and Yi Zhou and Ming-Hsuan Yang and Zhixin Shu},
      year={2024},
      eprint={2412.17812},
      archivePrefix={arXiv},
      primaryClass={cs.CV}
}

@inproceedings{corr2025lam,
  author       = {Yisheng He and
                  Xiaodong Gu and
                  Xiaodan Ye and
                  Chao Xu and
                  Zhengyi Zhao and
                  Yuan Dong and
                  Weihao Yuan and
                  Zilong Dong and
                  Liefeng Bo},
  editor       = {Ginger Alford and
                  Hao (Richard) Zhang and
                  Adriana Schulz},
  title        = {{LAM:} Large Avatar Model for One-shot Animatable Gaussian Head},
  booktitle    = {Proceedings of the Special Interest Group on Computer Graphics and
                  Interactive Techniques Conference, {SIGGRAPH} Conference Papers 2025,
                  Vancouver, BC, Canada, August 10-14, 2025},
  pages        = {27:1--27:13},
  publisher    = {{ACM}},
  year         = {2025},
  url          = {https://doi.org/10.1145/3721238.3730706},
  doi          = {10.1145/3721238.3730706},
  bibsource    = {dblp computer science bibliography, https://dblp.org}
}

@article{corr2025lhm,
  author       = {Lingteng Qiu and
                  Xiaodong Gu and
                  Peihao Li and
                  Qi Zuo and
                  Weichao Shen and
                  Junfei Zhang and
                  Kejie Qiu and
                  Weihao Yuan and
                  Guanying Chen and
                  Zilong Dong and
                  Liefeng Bo},
  title        = {{LHM:} Large Animatable Human Reconstruction Model from a Single Image
                  in Seconds},
  journal      = {CoRR},
  volume       = {abs/2503.10625},
  year         = {2025},
  url          = {https://doi.org/10.48550/arXiv.2503.10625},
  doi          = {10.48550/ARXIV.2503.10625},
  eprinttype    = {arXiv},
  eprint       = {2503.10625},
  bibsource    = {dblp computer science bibliography, https://dblp.org}
}

@article{corr2025avat3r,
  author       = {Tobias Kirschstein and
                  Javier Romero and
                  Artem Sevastopolsky and
                  Matthias Nie{\ss}ner and
                  Shunsuke Saito},
  title        = {Avat3r: Large Animatable Gaussian Reconstruction Model for High-fidelity
                  3D Head Avatars},
  journal      = {CoRR},
  volume       = {abs/2502.20220},
  year         = {2025},
  url          = {https://doi.org/10.48550/arXiv.2502.20220},
  doi          = {10.48550/ARXIV.2502.20220},
  eprinttype    = {arXiv},
  eprint       = {2502.20220},
  bibsource    = {dblp computer science bibliography, https://dblp.org}
}

@inproceedings{cvpr2024dust3r,
  author       = {Shuzhe Wang and
                  Vincent Leroy and
                  Yohann Cabon and
                  Boris Chidlovskii and
                  J{\'{e}}r{\^{o}}me Revaud},
  title        = {DUSt3R: Geometric 3D Vision Made Easy},
  booktitle    = {{IEEE/CVF} Conference on Computer Vision and Pattern Recognition,
                  {CVPR} 2024, Seattle, WA, USA, June 16-22, 2024},
  pages        = {20697--20709},
  publisher    = {{IEEE}},
  year         = {2024},
  url          = {https://doi.org/10.1109/CVPR52733.2024.01956},
  doi          = {10.1109/CVPR52733.2024.01956},
  bibsource    = {dblp computer science bibliography, https://dblp.org}
}

@inproceedings{iclr2024lrm,
  author       = {Yicong Hong and
                  Kai Zhang and
                  Jiuxiang Gu and
                  Sai Bi and
                  Yang Zhou and
                  Difan Liu and
                  Feng Liu and
                  Kalyan Sunkavalli and
                  Trung Bui and
                  Hao Tan},
  title        = {{LRM:} Large Reconstruction Model for Single Image to 3D},
  booktitle    = {The Twelfth International Conference on Learning Representations,
                  {ICLR} 2024, Vienna, Austria, May 7-11, 2024},
  publisher    = {OpenReview.net},
  year         = {2024},
  url          = {https://openreview.net/forum?id=sllU8vvsFF},
  bibsource    = {dblp computer science bibliography, https://dblp.org}
}

@inproceedings{eccv2024lgm,
  author       = {Jiaxiang Tang and
                  Zhaoxi Chen and
                  Xiaokang Chen and
                  Tengfei Wang and
                  Gang Zeng and
                  Ziwei Liu},
  editor       = {Ales Leonardis and
                  Elisa Ricci and
                  Stefan Roth and
                  Olga Russakovsky and
                  Torsten Sattler and
                  G{\"{u}}l Varol},
  title        = {{LGM:} Large Multi-view Gaussian Model for High-Resolution 3D Content
                  Creation},
  booktitle    = {Computer Vision - {ECCV} 2024 - 18th European Conference, Milan, Italy,
                  September 29-October 4, 2024, Proceedings, Part {IV}},
  series       = {Lecture Notes in Computer Science},
  volume       = {15062},
  pages        = {1--18},
  publisher    = {Springer},
  year         = {2024},
  url          = {https://doi.org/10.1007/978-3-031-73235-5\_1},
  doi          = {10.1007/978-3-031-73235-5\_1},
  bibsource    = {dblp computer science bibliography, https://dblp.org}
}

@inproceedings{eg2000real-time-hair,
  author       = {C. K. Koh and
                  Z. Huang},
  editor       = {Nadia Magnenat{-}Thalmann and
                  Daniel Thalmann and
                  Bruno Arnaldi},
  title        = {Real-Time Animation of Human Hair Modeled in Strips},
  booktitle    = {Proceedings of the Eurographics Workshop on Computer Animation and
                  Simulation 2000, Interlaken, Switzerland, August 21-22, 2000},
  series       = {Eurographics},
  pages        = {101--110},
  publisher    = {Springer},
  year         = {2000},
  url          = {https://doi.org/10.1007/978-3-7091-6344-3\_8},
  doi          = {10.1007/978-3-7091-6344-3\_8},
  bibsource    = {dblp computer science bibliography, https://dblp.org}
}

@inproceedings{pg2003hair-animation,
  author       = {Wenqi Liang and
                  Zhiyong Huang},
  title        = {An Enhanced Framework for Real-Time Hair Animation},
  booktitle    = {11th Pacific Conference on Computer Graphics and Applications, {PG}
                  2003, Canmore, Canada, October 8-10, 2003},
  pages        = {467--471},
  publisher    = {{IEEE} Computer Society},
  year         = {2003},
  url          = {https://doi.org/10.1109/PCCGA.2003.1238296},
  doi          = {10.1109/PCCGA.2003.1238296},
  bibsource    = {dblp computer science bibliography, https://dblp.org}
}

@inproceedings{cgi2004hair-nurbs,
  author       = {Paul Noble and
                  Wen Tang},
  title        = {Modelling and Animating Cartoon Hair with {NURBS} Surfaces},
  booktitle    = {2004 Computer Graphics International {(CGI} 2004), 16-19 June 2004,
                  Crete, Greece},
  pages        = {60--67},
  publisher    = {{IEEE} Computer Society},
  year         = {2004},
  url          = {https://doi.org/10.1109/CGI.2004.1309193},
  doi          = {10.1109/CGI.2004.1309193},
  bibsource    = {dblp computer science bibliography, https://dblp.org}
}

@article{tog2004hair-multi,
  author       = {Sylvain Paris and
                  H{\'{e}}ctor M. Brice{\~{n}}o and
                  Fran{\c{c}}ois X. Sillion},
  title        = {Capture of hair geometry from multiple images},
  journal      = {{ACM} Trans. Graph.},
  volume       = {23},
  number       = {3},
  pages        = {712--719},
  year         = {2004},
  url          = {https://doi.org/10.1145/1015706.1015784},
  doi          = {10.1145/1015706.1015784},
  bibsource    = {dblp computer science bibliography, https://dblp.org}
}

@inproceedings{cvpr2019strand-mvs,
  author       = {Giljoo Nam and
                  Chenglei Wu and
                  Min H. Kim and
                  Yaser Sheikh},
  title        = {Strand-Accurate Multi-View Hair Capture},
  booktitle    = {{IEEE} Conference on Computer Vision and Pattern Recognition, {CVPR}
                  2019, Long Beach, CA, USA, June 16-20, 2019},
  pages        = {155--164},
  publisher    = {Computer Vision Foundation / {IEEE}},
  year         = {2019},
  url          = {http://openaccess.thecvf.com/content\_CVPR\_2019/html/Nam\_Strand-Accurate\_Multi-View\_Hair\_Capture\_CVPR\_2019\_paper.html},
  doi          = {10.1109/CVPR.2019.00024},
  bibsource    = {dblp computer science bibliography, https://dblp.org}
}

@inproceedings{eg2021hair-inverse,
  author       = {Tiancheng Sun and
                  Giljoo Nam and
                  Carlos Aliaga and
                  Christophe Hery and
                  Ravi Ramamoorthi},
  editor       = {Adrien Bousseau and
                  Morgan McGuire},
  title        = {Human Hair Inverse Rendering using Multi-View Photometric data},
  booktitle    = {32nd Eurographics Symposium on Rendering, {EGSR} 2021 - Digital Library
                  Only Track, Saarbr{\"{u}}cken, Germany, June 29 - July 2, 2021},
  pages        = {179--190},
  publisher    = {Eurographics Association},
  year         = {2021},
  url          = {https://doi.org/10.2312/sr.20211301},
  doi          = {10.2312/SR.20211301},
  bibsource    = {dblp computer science bibliography, https://dblp.org}
}

@inproceedings{eccv2022neural-strands,
  author       = {Radu Alexandru Rosu and
                  Shunsuke Saito and
                  Ziyan Wang and
                  Chenglei Wu and
                  Sven Behnke and
                  Giljoo Nam},
  editor       = {Shai Avidan and
                  Gabriel J. Brostow and
                  Moustapha Ciss{\'{e}} and
                  Giovanni Maria Farinella and
                  Tal Hassner},
  title        = {Neural Strands: Learning Hair Geometry and Appearance from Multi-view
                  Images},
  booktitle    = {Computer Vision - {ECCV} 2022 - 17th European Conference, Tel Aviv,
                  Israel, October 23-27, 2022, Proceedings, Part {XXXIII}},
  series       = {Lecture Notes in Computer Science},
  volume       = {13693},
  pages        = {73--89},
  publisher    = {Springer},
  year         = {2022},
  url          = {https://doi.org/10.1007/978-3-031-19827-4\_5},
  doi          = {10.1007/978-3-031-19827-4\_5},
  bibsource    = {dblp computer science bibliography, https://dblp.org}
}

@inproceedings{iccv2023neural-haircut,
  author       = {Vanessa Sklyarova and
                  Jenya Chelishev and
                  Andreea Dogaru and
                  Igor Medvedev and
                  Victor Lempitsky and
                  Egor Zakharov},
  title        = {Neural Haircut: Prior-Guided Strand-Based Hair Reconstruction},
  booktitle    = {{IEEE/CVF} International Conference on Computer Vision, {ICCV} 2023,
                  Paris, France, October 1-6, 2023},
  pages        = {19705--19716},
  publisher    = {{IEEE}},
  year         = {2023},
  url          = {https://doi.org/10.1109/ICCV51070.2023.01810},
  doi          = {10.1109/ICCV51070.2023.01810},
  bibsource    = {dblp computer science bibliography, https://dblp.org}
}

@inproceedings{cvpr2023neuwigs,
  author       = {Ziyan Wang and
                  Giljoo Nam and
                  Tuur Stuyck and
                  Stephen Lombardi and
                  Chen Cao and
                  Jason M. Saragih and
                  Michael Zollh{\"{o}}fer and
                  Jessica K. Hodgins and
                  Christoph Lassner},
  title        = {NeuWigs: {A} Neural Dynamic Model for Volumetric Hair Capture and
                  Animation},
  booktitle    = {{IEEE/CVF} Conference on Computer Vision and Pattern Recognition,
                  {CVPR} 2023, Vancouver, BC, Canada, June 17-24, 2023},
  pages        = {8641--8651},
  publisher    = {{IEEE}},
  year         = {2023},
  url          = {https://doi.org/10.1109/CVPR52729.2023.00835},
  doi          = {10.1109/CVPR52729.2023.00835},
  bibsource    = {dblp computer science bibliography, https://dblp.org}
}

@article{corr2023delta,
  author       = {Yao Feng and
                  Weiyang Liu and
                  Timo Bolkart and
                  Jinlong Yang and
                  Marc Pollefeys and
                  Michael J. Black},
  title        = {Learning Disentangled Avatars with Hybrid 3D Representations},
  journal      = {CoRR},
  volume       = {abs/2309.06441},
  year         = {2023},
  url          = {https://doi.org/10.48550/arXiv.2309.06441},
  doi          = {10.48550/ARXIV.2309.06441},
  eprinttype    = {arXiv},
  eprint       = {2309.06441},
  bibsource    = {dblp computer science bibliography, https://dblp.org}
}

@article{tog2018hair-vae,
  author       = {Shunsuke Saito and
                  Liwen Hu and
                  Chongyang Ma and
                  Hikaru Ibayashi and
                  Linjie Luo and
                  Hao Li},
  title        = {3D hair synthesis using volumetric variational autoencoders},
  journal      = {{ACM} Trans. Graph.},
  volume       = {37},
  number       = {6},
  pages        = {208},
  year         = {2018},
  url          = {https://doi.org/10.1145/3272127.3275019},
  doi          = {10.1145/3272127.3275019},
  bibsource    = {dblp computer science bibliography, https://dblp.org}
}

@inproceedings{3dv2024teca,
  author       = {Hao Zhang and
                  Yao Feng and
                  Peter Kulits and
                  Yandong Wen and
                  Justus Thies and
                  Michael J. Black},
  title        = {{TECA:} Text-Guided Generation and Editing of Compositional 3D Avatars},
  booktitle    = {International Conference on 3D Vision, 3DV 2024, Davos, Switzerland,
                  March 18-21, 2024},
  pages        = {1520--1530},
  publisher    = {{IEEE}},
  year         = {2024},
  url          = {https://doi.org/10.1109/3DV62453.2024.00151},
  doi          = {10.1109/3DV62453.2024.00151},
  bibsource    = {dblp computer science bibliography, https://dblp.org}
}

@article{tog2009hair-meshes,
  author       = {Cem Yuksel and
                  Scott Schaefer and
                  John Keyser},
  title        = {Hair meshes},
  journal      = {{ACM} Trans. Graph.},
  volume       = {28},
  number       = {5},
  pages        = {166},
  year         = {2009},
  url          = {https://doi.org/10.1145/1618452.1618512},
  doi          = {10.1145/1618452.1618512},
  bibsource    = {dblp computer science bibliography, https://dblp.org}
}

@article{tog20253dgh,
  author       = {Chengan He and
                  Junxuan Li and
                  Tobias Kirschstein and
                  Artem Sevastopolsky and
                  Shunsuke Saito and
                  Qingyang Tan and
                  Javier Romero and
                  Chen Cao and
                  Holly E. Rushmeier and
                  Giljoo Nam},
  title        = {3DGH: 3D Head Generation with Composable Hair and Face},
  journal      = {{ACM} Trans. Graph.},
  volume       = {44},
  number       = {4},
  pages        = {59:1--59:12},
  year         = {2025},
  url          = {https://doi.org/10.1145/3731211},
  doi          = {10.1145/3731211},
  bibsource    = {dblp computer science bibliography, https://dblp.org}
}

@misc{corr2024hhavatar,
      title={HHAvatar: Gaussian Head Avatar with Dynamic Hairs}, 
      author={Zhanfeng Liao and Yuelang Xu and Zhe Li and Qijing Li and Boyao Zhou and Ruifeng Bai and Di Xu and Hongwen Zhang and Yebin Liu},
      year={2024},
      eprint={2312.03029},
      archivePrefix={arXiv},
      primaryClass={cs.CV},
      url={https://arxiv.org/abs/2312.03029}, 
}

@misc{corr2025hairgs,
      title={HairGS: Hair Strand Reconstruction based on 3D Gaussian Splatting}, 
      author={Yimin Pan and Matthias Nießner and Tobias Kirschstein},
      year={2025},
      eprint={2509.07774},
      archivePrefix={arXiv},
      primaryClass={cs.CV},
      url={https://arxiv.org/abs/2509.07774}, 
}

@inproceedings{eccv2024gaussian-haircut,
  author       = {Egor Zakharov and
                  Vanessa Sklyarova and
                  Michael J. Black and
                  Giljoo Nam and
                  Justus Thies and
                  Otmar Hilliges},
  editor       = {Ales Leonardis and
                  Elisa Ricci and
                  Stefan Roth and
                  Olga Russakovsky and
                  Torsten Sattler and
                  G{\"{u}}l Varol},
  title        = {Human Hair Reconstruction with Strand-Aligned 3D Gaussians},
  booktitle    = {Computer Vision - {ECCV} 2024 - 18th European Conference, Milan, Italy,
                  September 29-October 4, 2024, Proceedings, Part {XVI}},
  series       = {Lecture Notes in Computer Science},
  volume       = {15074},
  pages        = {409--425},
  publisher    = {Springer},
  year         = {2024},
  url          = {https://doi.org/10.1007/978-3-031-72640-8\_23},
  doi          = {10.1007/978-3-031-72640-8\_23},
  bibsource    = {dblp computer science bibliography, https://dblp.org}
}

@article{corr2024gaussian-hair,
  author       = {Haimin Luo and
                  Min Ouyang and
                  Zijun Zhao and
                  Suyi Jiang and
                  Longwen Zhang and
                  Qixuan Zhang and
                  Wei Yang and
                  Lan Xu and
                  Jingyi Yu},
  title        = {GaussianHair: Hair Modeling and Rendering with Light-aware Gaussians},
  journal      = {CoRR},
  volume       = {abs/2402.10483},
  year         = {2024},
  url          = {https://doi.org/10.48550/arXiv.2402.10483},
  doi          = {10.48550/ARXIV.2402.10483},
  eprinttype    = {arXiv},
  eprint       = {2402.10483},
  bibsource    = {dblp computer science bibliography, https://dblp.org}
}

@inproceedings{iclr2025perm,
  author       = {Chengan He and
                  Xin Sun and
                  Zhixin Shu and
                  Fujun Luan and
                  S{\"{o}}ren Pirk and
                  Jorge Alejandro Amador Herrera and
                  Dominik Ludewig Michels and
                  Tuanfeng Yang Wang and
                  Meng Zhang and
                  Holly E. Rushmeier and
                  Yi Zhou},
  title        = {Perm: {A} Parametric Representation for Multi-Style 3D Hair Modeling},
  booktitle    = {The Thirteenth International Conference on Learning Representations,
                  {ICLR} 2025, Singapore, April 24-28, 2025},
  publisher    = {OpenReview.net},
  year         = {2025},
  url          = {https://openreview.net/forum?id=WKfb1xGXGx},
  bibsource    = {dblp computer science bibliography, https://dblp.org}
}

@inproceedings{cvpr2023diffusionrig,
  author       = {Zheng Ding and
                  Xuaner Zhang and
                  Zhihao Xia and
                  Lars Jebe and
                  Zhuowen Tu and
                  Xiuming Zhang},
  title        = {DiffusionRig: Learning Personalized Priors for Facial Appearance Editing},
  booktitle    = {{IEEE/CVF} Conference on Computer Vision and Pattern Recognition,
                  {CVPR} 2023, Vancouver, BC, Canada, June 17-24, 2023},
  pages        = {12736--12746},
  publisher    = {{IEEE}},
  year         = {2023},
  url          = {https://doi.org/10.1109/CVPR52729.2023.01225},
  doi          = {10.1109/CVPR52729.2023.01225},
  bibsource    = {dblp computer science bibliography, https://dblp.org}
}

@inproceedings{cvpr2019arcface,
  author       = {Jiankang Deng and
                  Jia Guo and
                  Niannan Xue and
                  Stefanos Zafeiriou},
  title        = {ArcFace: Additive Angular Margin Loss for Deep Face Recognition},
  booktitle    = {{IEEE} Conference on Computer Vision and Pattern Recognition, {CVPR}
                  2019, Long Beach, CA, USA, June 16-20, 2019},
  pages        = {4690--4699},
  publisher    = {Computer Vision Foundation / {IEEE}},
  year         = {2019},
  url          = {http://openaccess.thecvf.com/content\_CVPR\_2019/html/Deng\_ArcFace\_Additive\_Angular\_Margin\_Loss\_for\_Deep\_Face\_Recognition\_CVPR\_2019\_paper.html},
  doi          = {10.1109/CVPR.2019.00482},
  bibsource    = {dblp computer science bibliography, https://dblp.org}
}

@article{ijcv2021pipnet,
  author       = {Haibo Jin and
                  Shengcai Liao and
                  Ling Shao},
  title        = {Pixel-in-Pixel Net: Towards Efficient Facial Landmark Detection in
                  the Wild},
  journal      = {Int. J. Comput. Vis.},
  volume       = {129},
  number       = {12},
  pages        = {3174--3194},
  year         = {2021},
  url          = {https://doi.org/10.1007/s11263-021-01521-4},
  doi          = {10.1007/S11263-021-01521-4},
  bibsource    = {dblp computer science bibliography, https://dblp.org}
}

@inproceedings{cvpr2025rgbavatar,
  author       = {Linzhou Li and
                  Yumeng Li and
                  Yanlin Weng and
                  Youyi Zheng and
                  Kun Zhou},
  title        = {RGBAvatar: Reduced Gaussian Blendshapes for Online Modeling of Head
                  Avatars},
  booktitle    = {{IEEE/CVF} Conference on Computer Vision and Pattern Recognition,
                  {CVPR} 2025, Nashville, TN, USA, June 11-15, 2025},
  pages        = {10747--10757},
  publisher    = {Computer Vision Foundation / {IEEE}},
  year         = {2025},
  url          = {https://openaccess.thecvf.com/content/CVPR2025/html/Li\_RGBAvatar\_Reduced\_Gaussian\_Blendshapes\_for\_Online\_Modeling\_of\_Head\_Avatars\_CVPR\_2025\_paper.html},
  doi          = {10.1109/CVPR52734.2025.01004},
  bibsource    = {dblp computer science bibliography, https://dblp.org}
}

@inproceedings{siggraph2024gghead,
  author       = {Tobias Kirschstein and
                  Simon Giebenhain and
                  Jiapeng Tang and
                  Markos Georgopoulos and
                  Matthias Nie{\ss}ner},
  editor       = {Takeo Igarashi and
                  Ariel Shamir and
                  Hao (Richard) Zhang},
  title        = {GGHead: Fast and Generalizable 3D Gaussian Heads},
  booktitle    = {{SIGGRAPH} Asia 2024 Conference Papers, {SA} 2024, Tokyo, Japan, December
                  3-6, 2024},
  pages        = {126:1--126:11},
  publisher    = {{ACM}},
  year         = {2024},
  url          = {https://doi.org/10.1145/3680528.3687686},
  doi          = {10.1145/3680528.3687686},
  bibsource    = {dblp computer science bibliography, https://dblp.org}
}

@inproceedings{icassp2025avatargan,
  author       = {Guohao Li and
                  Hongyu Yang and
                  Yifang Men and
                  Di Huang and
                  Weixin Li and
                  Ruijie Yang and
                  Yunhong Wang},
  title        = {Generating Editable Head Avatars with 3D Gaussian GANs},
  booktitle    = {2025 {IEEE} International Conference on Acoustics, Speech and Signal
                  Processing, {ICASSP} 2025, Hyderabad, India, April 6-11, 2025},
  pages        = {1--5},
  publisher    = {{IEEE}},
  year         = {2025},
  url          = {https://doi.org/10.1109/ICASSP49660.2025.10887885},
  doi          = {10.1109/ICASSP49660.2025.10887885},
  bibsource    = {dblp computer science bibliography, https://dblp.org}
}

@article{corr2025pi3,
  author       = {Yifan Wang and
                  Jianjun Zhou and
                  Haoyi Zhu and
                  Wenzheng Chang and
                  Yang Zhou and
                  Zizun Li and
                  Junyi Chen and
                  Jiangmiao Pang and
                  Chunhua Shen and
                  Tong He},
  title        = {{\(\pi\)}\({}^{\mbox{3}}\): Scalable Permutation-Equivariant Visual
                  Geometry Learning},
  journal      = {CoRR},
  volume       = {abs/2507.13347},
  year         = {2025},
  url          = {https://doi.org/10.48550/arXiv.2507.13347},
  doi          = {10.48550/ARXIV.2507.13347},
  eprinttype    = {arXiv},
  eprint       = {2507.13347},
  bibsource    = {dblp computer science bibliography, https://dblp.org}
}

@inproceedings{cvpr2025perse,
  author       = {Hyunsoo Cha and
                  Inhee Lee and
                  Hanbyul Joo},
  title        = {{PERSE:} Personalized 3D Generative Avatars from {A} Single Portrait},
  booktitle    = {{IEEE/CVF} Conference on Computer Vision and Pattern Recognition,
                  {CVPR} 2025, Nashville, TN, USA, June 11-15, 2025},
  pages        = {15953--15962},
  publisher    = {Computer Vision Foundation / {IEEE}},
  year         = {2025},
  url          = {https://openaccess.thecvf.com/content/CVPR2025/html/Cha\_PERSE\_Personalized\_3D\_Generative\_Avatars\_from\_A\_Single\_Portrait\_CVPR\_2025\_paper.html},
  doi          = {10.1109/CVPR52734.2025.01487},
  bibsource    = {dblp computer science bibliography, https://dblp.org}
}

@inproceedings{cvpr2024spalttingavatar,
  author       = {Zhijing Shao and
                  Zhaolong Wang and
                  Zhuang Li and
                  Duotun Wang and
                  Xiangru Lin and
                  Yu Zhang and
                  Mingming Fan and
                  Zeyu Wang},
  title        = {SplattingAvatar: Realistic Real-Time Human Avatars With Mesh-Embedded
                  Gaussian Splatting},
  booktitle    = {{IEEE/CVF} Conference on Computer Vision and Pattern Recognition,
                  {CVPR} 2024, Seattle, WA, USA, June 16-22, 2024},
  pages        = {1606--1616},
  publisher    = {{IEEE}},
  year         = {2024},
  url          = {https://doi.org/10.1109/CVPR52733.2024.00159},
  doi          = {10.1109/CVPR52733.2024.00159},
  bibsource    = {dblp computer science bibliography, https://dblp.org}
}

@inproceedings{cvpr2025mega,
  author       = {Cong Wang and
                  Di Kang and
                  Heyi Sun and
                  Shen{-}Han Qian and
                  Zixuan Wang and
                  Linchao Bao and
                  Song{-}Hai Zhang},
  title        = {MeGA: Hybrid Mesh-Gaussian Head Avatar for High-Fidelity Rendering
                  and Head Editing},
  booktitle    = {{IEEE/CVF} Conference on Computer Vision and Pattern Recognition,
                  {CVPR} 2025, Nashville, TN, USA, June 11-15, 2025},
  pages        = {26274--26284},
  publisher    = {Computer Vision Foundation / {IEEE}},
  year         = {2025},
  url          = {https://openaccess.thecvf.com/content/CVPR2025/html/Wang\_MeGA\_Hybrid\_Mesh-Gaussian\_Head\_Avatar\_for\_High-Fidelity\_Rendering\_and\_Head\_CVPR\_2025\_paper.html},
  doi          = {10.1109/CVPR52734.2025.02447},
  bibsource    = {dblp computer science bibliography, https://dblp.org}
}

@inproceedings{cvpr2023pointavatar,
  author       = {Yufeng Zheng and
                  Wang Yifan and
                  Gordon Wetzstein and
                  Michael J. Black and
                  Otmar Hilliges},
  title        = {PointAvatar: Deformable Point-Based Head Avatars from Videos},
  booktitle    = {{IEEE/CVF} Conference on Computer Vision and Pattern Recognition,
                  {CVPR} 2023, Vancouver, BC, Canada, June 17-24, 2023},
  pages        = {21057--21067},
  publisher    = {{IEEE}},
  year         = {2023},
  url          = {https://doi.org/10.1109/CVPR52729.2023.02017},
  doi          = {10.1109/CVPR52729.2023.02017},
  bibsource    = {dblp computer science bibliography, https://dblp.org}
}
